\documentclass[journal]{IEEEtran}

\usepackage{cite}

\ifCLASSINFOpdf
   \usepackage[pdftex]{graphicx}
\else
\fi
\usepackage{amsmath}
\ifCLASSOPTIONcompsoc
  \usepackage[caption=false,font=normalsize,labelfont=sf,textfont=sf]{subfig}
\else
  \usepackage[caption=false,font=footnotesize]{subfig}
\fi
\usepackage{url}

\usepackage{bm}
\usepackage{tikz}
\usetikzlibrary{shapes,arrows,positioning,calc,backgrounds}
\usepackage{amssymb}
\usetikzlibrary{shapes,arrows,positioning,calc}
\usepackage{makecell}
\usepackage[bottom]{footmisc}
\usepackage{pifont}
\usepackage{color, colortbl}
\usepackage{multirow}
\usepackage{siunitx} 

\newcommand{\vect}[1]{\mathbf{#1}}

\newcommand{\cmark}{\ding{51} }

\definecolor{lightgray}{rgb}{0.83, 0.83, 0.83}

\DeclareMathOperator*{\argmax}{arg\,max}

\newcommand{\nlenv}[1]{\begin{tabular}{c} 
    #1
  \end{tabular}
}

\usepackage{xcolor}

\begin{document}

\title{Active Sensing for Search and Tracking: A Review}

\author{Luca~Varotto$^{1}$,
        Angelo~Cenedese$^{1}$,
        and~Andrea~Cavallaro$^{2}$%
\thanks{$^{1}$L. Varotto and A. Cenedese are with the Department of Information Engineering, University of Padova, Italy.
          Corresponding author: {\tt\small luca.varotto.5@phd.unipd.it}.}%
\thanks{$^{2}$A. Cavallaro is with the Centre for Intelligent Sensing, Queen Mary University of London, U.K.}
\thanks{This work was partially supported by the Department of Information Engineering under the BIRD-SEED TSTARK project and by the Gini Foundation of the University of Padova.}
}%

\markboth{IEEE TRANSACTIONS ON CYBERNETICS, VOL. xx, NO. yy, ...}%
{Shell \MakeLowercase{\textit{et al.}}: Bare Demo of IEEEtran.cls for IEEE Journals}
%



\maketitle


\begin{abstract}
Active Position Estimation (APE) is the task of localizing one or more targets using one or more sensing platforms. APE is a key task for search and rescue missions, wildlife monitoring, source term estimation, and collaborative mobile robotics. Success in APE depends on the level of cooperation of the sensing platforms, their number, their degrees of freedom and the quality of the information gathered.
APE control laws enable active sensing by satisfying either pure-exploitative or pure-explorative criteria. The former  minimizes the uncertainty on position estimation; whereas the latter drives the  platform closer to its task completion. 
In this paper, we define the main elements of APE  to systematically classify and critically discuss the state of the art in this domain. We also propose a reference framework as a formalism to classify APE-related solutions. 
Overall, this survey explores the principal challenges and
envisages the main research directions in the field of autonomous perception systems for localization tasks. It is also beneficial to promote the development of robust active sensing methods for search and tracking applications.
\end{abstract}


%
\IEEEpeerreviewmaketitle


\section{Introduction}\label{sec:intro}

\IEEEPARstart{A}{ctive} sensing
consists~\cite{aoki2011near} in the control of a dynamical system with actuation and sensing capabilities 
to automate the perception process and maximize its efficiency, in terms of acquired information or task completion~\cite{ryan2010particle,theoretical_sensor_management,active_sensing,bajcsy2018revisiting,bajcsy1988active}. By leveraging the interplay among estimation, perception, and control, active sensing is at the nexus between automation  and  robotics  research~\cite{parker2016multiple}. Fostered  by  the  higher performance guarantees compared to passive approaches~\cite{realTime_active_vision,active_classification} and enabling autonomous perception in robotics systems,
active sensing is employed in Search and Rescue (SAR) missions~\cite{SAR_radio,SAR_evolutionary,SAR_UAV}, environmental mapping~\cite{popovic2019enviromental,masehian2017cooperative} and exploration~\cite{ergodic_exploration,ghassemi2020extended,BayOpt_exploration}, wildlife monitoring~\cite{robotEcology,zebra_tracking,small_birds}, optimal sensor coverage~\cite{fabris2019distributed}, autonomous source term estimation (STE)~\cite{hutchinson2019unmanned,STE_review,park2020cooperative,ristic2016study,ristic2017autonomous,masson2009chasing,bourne2019coordinated}, object recognition, classification and manipulation~\cite{active_classification,calli2018active}, and collaborative mobile robotics~\cite{nagaty2015probabilistic,haugen2015monitoring,carron2015multi,yu2018algorithms,recharge}.  

{Active Position Estimation} (APE) is a specific active sensing task where one~\cite{DeepRL_gazeControl} or more~\cite{intermittent2009} platforms monitor and localize one~\cite{shahidian2017single} or more~\cite{chung2009coordinated} static~\cite{fink2010online} or dynamic~\cite{ryan2010particle} targets (humans~\cite{SAR_radio}, animals~\cite{robotEcology}, or robots~\cite{koohifar2018autonomous}), which may be active~\cite{gradientBased2012} or passive~\cite{liu2017model}.
%


Actuated sensing platforms drive the perception process towards observations that are informative optimal~\cite{shahidian2017single} or efficient in terms of task completion~\cite{DeepRL_gazeControl}. In the APE framework, information-seeking strategies~\cite{theoretical_sensor_management} adopt explorative policies that consist of making decisions for next actions in order to maximize the uncertainty reduction on the target position; this leads to better estimation performance with respect to passive sensor systems~\cite{information_focalLength,queralta2020collaborative}. On the contrary, task-driven approaches bring the sensing platform closer to its task execution, such as reaching a goal position within certain time/energy tolerances~\cite{task_drivenVSinformation_seeking}.

While a number of reviews on related topics exist~\cite{search_pursuit_review,cooperative_RinnerCavallaro,SAT_survey,detection_tracking_survey,bajcsy2018revisiting,stone1976theory,chen2011active,queralta2020collaborative}, we are the first to provide a formal definition of the APE problem (see Sec.~\ref{sec:APE}) and a comparative analysis on the main sensing modalities involved (see Sec.~\ref{sec:sensing}). 
The insight on the perception systems allows to highlight the role of information fusion and multi-agent cooperation in APE tasks (see Sec.~\ref{sec:info_fusion}).
Even though several works discuss the level of cooperation between sensing robots~\cite{cooperative_RinnerCavallaro,detection_tracking_survey,SAT_survey,queralta2020collaborative}, the importance of multi-modal fusion algorithms is mentioned only in~\cite{queralta2020collaborative}, which however focuses on SAR applications only. Our review generalizes to the entire APE class, which covers a broader spectrum of applications. 
As further contribution, we classify the state of the art according to the platforms degrees of freedom and the control methods adopted for the platform actuation (see Sec.~\ref{sec:dynamics}).
Finally, no previous analysis covers the main criteria that guide APE tasks; in this work, we classify the reviewed articles based on their level of exploration-exploitation trade-off, namely task-driven (pure exploitative) and information-seeking (pure explorative) control strategies (see Sec.~\ref{subsec:criteria}).
Fig.~\ref{fig:APE_scheme} illustrates the structure of this paper and a generic APE scheme.
\section{Active Position Estimation}\label{sec:APE}

\tikzset{
block/.style = {draw, fill=white, rectangle, minimum height=3em, minimum width=3em},
block_transp/.style = {rectangle, minimum height=3em, minimum width=3em},
sum/.style= {draw, fill=white, circle, node distance=1cm},
block_split_horiz/.style = {draw, fill=white,minimum height=3em, minimum width=3em,rectangle split, rectangle split horizontal, rectangle split parts=3},
block_split_vert/.style = {draw, fill=white,minimum height=3em, minimum width=3em,rectangle split, rectangle split parts=3}}
\begin{figure*}[t!]
\center
\resizebox{\textwidth}{!}{%
\begin{tikzpicture}[auto, node distance=3cm,>=latex',scale=0.7, transform shape]
\node [block] (g) {$f(\vect{p}_{t-1,\ell},\bm{\omega}_{t-1})$};
\node [block_transp, left of = g, xshift=0.8cm] (omega) { $\bm{\omega}_{t-1}$};      
\node [block, below of = g, yshift=1.5cm] (time_update_g) { $z^{-1}$};
\node [block, right of =g,xshift=5cm] (sensing) {
\textit{Sensing units}};
\node [block_split_horiz, below of = sensing,xshift=0.12cm] (likelihood) {
\nodepart{one}
$p(\vect{z}_{t,m}^{(1)}|\vect{p}_{t,\ell},\vect{s}_{t,m})$
\nodepart{two}
$\dots$
\nodepart{three}
$p(\vect{z}_{t,m}^{(C_m)}|\vect{p}_{t,\ell},\vect{s}_{t,m})$
};
\node [block, below of = likelihood,xshift=-0.1cm] (fusion) { \textit{Information fusion }};
\node [block, right of = fusion,xshift=3.5cm] (map) { \textit{Map update}};
\node [block, right of = map,xshift=1cm,yshift=1cm,minimum width=7em] (control) { \textit{Control}};
\node [block, above of = control,yshift=2cm] (sensor dynamics) { $\vect{s}_{t,m} = g_m\left( \vect{s}_{t-1,m},\vect{u}_{t-1}^*\right)$};
\node [block, below of = sensor dynamics, yshift=1.2cm,xshift=1cm] (time_update_sensor) { $z^{-1}$};
\node [block_split_vert, left of = fusion,xshift=-5cm] (other_platforms) {
\nodepart{one}
$1$-st platform
\nodepart{two}
$\dots$
\nodepart{three}
$M$-th platform
};

\draw [->] (omega.east) --  node{} (g.west);
\draw [->] (g.east) -- node[]{ $\vect{p}_{t,\ell}$} (sensing.west);
\draw [->] ($(g.east)+(0.5,0)$) |- (time_update_g.east) ;
\draw [->] (time_update_g.west) -- ++ (-1,0) -- node[]{ $\vect{p}_{t-1,\ell}$} ++ (0,1.2) -- ($(g.west)-(0,0.3)$);
\draw [->] ($(g)+(3,0)$) |- ($(other_platforms.east)+(0.3,0.8)$);
\draw [->] (sensor dynamics.west) -- node[xshift=1.5cm]{$\vect{s}_{t,m}$} (sensing);
\draw [->] ($(sensor dynamics.west)+(-0.2,0)$) |- (time_update_sensor.west) ;
\draw [->] (time_update_sensor.east) -- ++ (0.7,0) -- node[]{ $\vect{s}_{t-1,m},\vect{u}_{t-1}^*$} ++ (0,1.8) -- ($(sensor dynamics.east)$);
\draw [->] ($(time_update_sensor.west) + (-0.5,0)$) -- ($(control.north)$) ;
\draw [->] ($(fusion.west) + (0,0.2)$) -- node[yshift=0.5cm,xshift=-0.5cm]{$\vect{z}_{t,m}$} ($(other_platforms)+(1.5,0.2)$) ;
\draw [->] ($(other_platforms) + (1.5,-0.2)$) -- node[yshift=-1cm,xshift=-0.5cm]{ \nlenv{$    \{\vect{z}_{t,m^\prime}\}_{m^\prime=1}^M$ \\ $m^\prime \neq m$}} ($(fusion.west)+(0,-0.2)$) ;
\draw [->] (sensing.south) -- node[xshift=-1cm,yshift=0.5cm]{$\vect{z}_{t,m}^{(1)}$} (likelihood.one north);
\draw [->] (sensing.south) -- node[]{$\dots$} (likelihood.two north);
\draw [->] (sensing.south) -- node[yshift=-0.2cm]{$\vect{z}_{t,m}^{(C_m)}$} (likelihood.three north);
\draw [->] (likelihood.one south) --  (fusion.north);
\draw [->] (likelihood.two south) --  (fusion.north);
\draw [->] (likelihood.three south) --  (fusion.north);
\draw [->] (fusion.east) --  node[]{$p(\vect{z}_{t}|\vect{p}_{t,\ell},\vect{s}_{t,1:M})$}(map.west);
\draw [->] (map.east) -|  node[yshift=-0.3cm,xshift=-0.5cm]{$p(\vect{p}_{t,\ell}|\vect{z}_{1:t})$}($(control.south)+(-0.2,0)$);
\draw [->] ($(control.north)+(1,0)$) -- node[]{ $\vect{u_t}^*$}(time_update_sensor.south);
\draw [->] (fusion.south) |- node[xshift=2cm]{ $\vect{s}_{t,1:M}$} ++ (9,-1) -|  ($(control.south)+(0.2,0)$);

\begin{scope}[on background layer]

\draw [dashed] ($(time_update_g) + (-3,-0.7)$) rectangle ($(time_update_g) + (2,2.2)$);
\draw [dashed,fill=lightgray] ($(likelihood.west) + (-0.2,-0.8)$) rectangle ($(sensing.east) + (4.5,0.8)$);
\draw [dashed,fill=lightgray] ($(sensor dynamics.west) + (-0.5,0.7)$) rectangle ($(control.east) + (1.2,-0.8)$);
\draw [dashed,fill=lightgray] ($(fusion.west) + (-0.2,0.8)$) rectangle ($(fusion.east) + (0.2,-0.7)$);
\draw [dashed,fill=lightgray] ($(map.west) + (-0.2,0.8)$) rectangle ($(map.east) + (0.2,-0.7)$);
\draw [dashed] ($(sensing) + (-3.5,1.4)$) rectangle ($(control) + (2.7,-2.7)$);
\draw [dashed] ($(other_platforms) + (-1.5,1)$) rectangle ($(other_platforms) + (1.5,-1)$);

\end{scope}

\node [block_transp, above of = g,yshift=-2cm] (target_annotation) { \textit{Target dynamics} ({\large \textbf{Sec. \ref{sec:APE}})}};
\node [block_transp, above of = sensor dynamics,yshift=-2cm] (sensor_annotation) { \textit{Platform dynamics} ({\large\textbf{Sec. \ref{sec:dynamics}-\ref{subsec:criteria}})}};
\node [block_transp, right of = likelihood,xshift=1.3cm] (map_annotation) {\nlenv{\textit{Observation} \\ \textit{models} }};
\node [block_transp, above of = sensing, yshift=-2cm] (sensing_annotation_sec) {{\large (\textbf{Sec. \ref{sec:sensing}})}};
\node [block_transp, below of = fusion,yshift=2cm,xshift=-1cm] (fusion_annotation_sec) { ({\large \textbf{Sec. \ref{sec:info_fusion}})}};
\node [block_transp, below of = map, yshift=2cm] (map_annotation) { \textit{Probabilistic map}};
\node [block_transp, above of = map, yshift=-2cm] (map_annotation_sec) { {\large (\textbf{Sec. \ref{sec:info_fusion}})}};
\node [block_transp, above of = sensor dynamics, yshift=-1.3cm] (sensor_annotation) { \textit{$m$-th sensing platform} 
({\large \textbf{Sec. \ref{sec:APE}})}
};
\end{tikzpicture}
}
\vspace{-0.2cm}
\caption{ Generic APE scheme: 
different targets properties require different APE algorithms and, therefore, sensing platforms with specific requirements (Sec. \ref{sec:APE}). Perception modules (Sec. \ref{sec:sensing}), on-board data management and multi-agent coordination (Sec. \ref{sec:info_fusion}), and platform control (Sec. \ref{sec:dynamics}-\ref{subsec:criteria}) are the most important aspects to be considered in the design of any APE mission. }
\label{fig:APE_scheme}
\end{figure*}
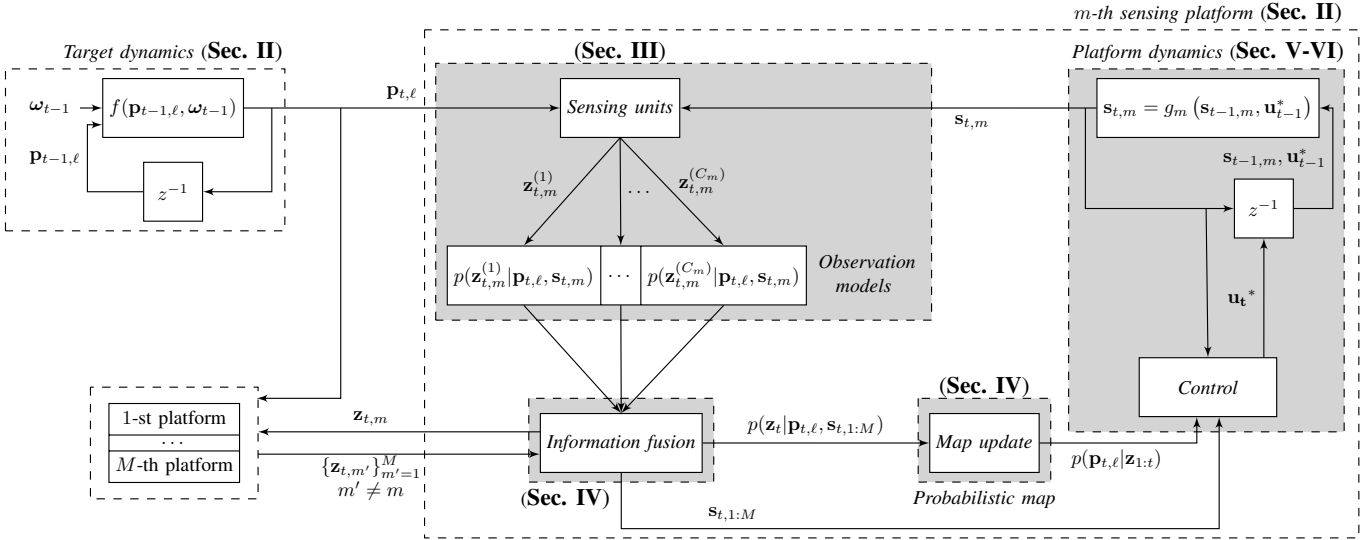

Robotic platforms represent a propulsive technology for data acquisition in civil and industrial contexts, as well as in a wide variety of applications including environmental monitoring and exploration, precision agriculture, surveillance, and post-disaster assessment~\cite{popovic2019enviromental}. Most robotic applications hinge on localization modules to accomplish specific tasks (e.g., multi-robot rendevouz~\cite{recharge}). In this framework, the main challenge is to design autonomous and cooperative decision making strategies to search and localize the targets within a set of given resource constraints (e.g., energy consumption, mission time, travel distance)~\cite{popovic2019enviromental}. For this reason, APE algorithms have become pervasive in a variety of applications, including those related to human-robot cooperation and mobile robotics.

An APE task can be categorized as Active Search, Localization and Tracking. Active Search refers to the problem of making data-collection decisions in order to search for objects of interest (i.e., targets) which have not been detected yet~\cite{ghods2020multi}. In Active Localization, the purpose is to estimate the location of static targets that have already been detected~\cite{hutchinson2019unmanned,fink2010online,park2020cooperative,gradientBased2012,shahidian2017single}; Active Tracking differs from Active Localization in that the targets are dynamic~\cite{wang2018tracking}. 
Furthermore, in a Search and Tracking task, Active Search precedes (or coexists with) Active Tracking~\cite{DeepRL_gazeControl,intermittent2009,liu2017model}. 
In this Section, we introduce the main elements of the APE scheme, namely the targets and the sensing platforms. We also provide a formal characterization of the APE classification system. 


Tab.~\ref{tab:summary_table} organizes the reviewed articles according to the APE method applied,
the characteristics of the targets,
as well as the number of sensing platforms involved.

\subsection{The targets}\label{subsec:targets}

Let $N$ be the number of targets to be searched and tracked\footnote{In practice, $N$ is not always known beforehand and it may also be time-varying. In these cases, the uncertainty of the APE problem increases; hence, ad-hoc and more complex planning algorithms are required~\cite{yi2012target}}. 
The movements of {\em target} $\ell$ at time instant $t$ can be described by a stochastic Markovian state transition model~\cite{optimal_information_search}:
\begin{equation}\label{eq:target_dynamics}
\begin{split}
    & \vect{p}_{t+1,\ell} = f_{\ell}(\vect{p}_{t,\ell},\bm{\omega}_{t,\ell}); \; \ell = 1, \dots, N\\
    & \text{s.t. } \vect{p}_{0,\ell} \sim \mathcal{P}_0 \\
    & \hspace{0.6cm} \vect{p}_{t,\ell} \in \Pi, \; t \geq 0 \\
    & \hspace{0.6cm} \bm{\omega}_{t,\ell} \sim \bm{\Omega}_{\ell}
\end{split}
\end{equation}
where $\vect{p}_{t,\ell}$ is the target position, $\vect{p}_{0,\ell}$ is the initial condition, realization of the random vector with probability density $\mathcal{P}_0$, $\Pi \subset \mathbb{R}^d$ ($d \leq 3$) is the workspace (i.e.~the environment where targets move); and  $\bm{\omega}_{t,\ell}$ is the process noise, realization of a random vector with probability density $\bm{\Omega}_{\ell}$ and support in $\mathbb{R}^d$. More specifically, the bias of $\bm{\Omega}_{\ell}$ (i.e., $\mathbb{E}[\bm{\omega}_{\ell}]$) models the target driving input, while the variance (i.e., $var\left[\bm{\omega}_{\ell}\right]$) accounts for the model inaccuracies and the state uncertainty. The process model is described by $p(\vect{p}_{t+1,\ell}|\vect{p}_{t,\ell})$, which is defined by \mbox{$f_{\ell}: \Pi \times \mathbb{R}^d \rightarrow \Pi$}, a (possibly) non-linear function that describes the dynamics of  target $\ell$~\cite{optimal_information_search}. Typically, \eqref{eq:target_dynamics} can be simplified by assuming the same dynamic behavior and process noise distribution for all targets~\cite{wong2005multi}, namely
\begin{equation}
    f_{\ell}(\cdot) = f(\cdot), \; \bm{\Omega}_{\ell} = \bm{\Omega}; \;  \ell=1,\dots,N.
\end{equation}
A typical choice for $\bm{\Omega}$ is the Normal distribution~\cite{optimal_information_search},
\begin{equation}
\bm{\Omega} = \mathcal{N}\left( \bm{\mu}_{\bm{\Omega}},\bm{\Sigma}_{\bm{\Omega}} \right).    
\end{equation}
where $\bm{\mu}_{\bm{\Omega}}$ and $\bm{\Sigma}_{\bm{\Omega}}$ are the bias and variance of $\bm{\Omega}$, respectively.

In the APE literature, a target may be {\em active} or {\em passive}~\cite{deak2012survey}. 

\begin{table}[b!]
\centering
\caption{Comparison of APE methods.
}
\label{tab:summary_table}
\begin{tabular}{c |c |lcr |r}

\hline
\multirow{2}{*}{\textbf{Ref.}} &
\multirow{2}{*}{\textbf{Class}} &
\multicolumn{3}{c|}{\textbf{Target}} &
\multirow{2}{*}{\makecell{{$M$}}}\\ 
\cline{3-5}
& & \multicolumn{1}{c}{{Motion}} & {Type}  & {$N$} & \\ \hline
\rowcolor{lightgray}
\cite{liu2015model} & SAT  & static  & P  & $1$ & $1$\\
\cite{liu2017model,ramirez2014moving} & SAT  & dynamic  & P  & $1$ & $1$\\ 
\rowcolor{lightgray}
\cite{DeepRL_gazeControl} & SAT & dynamic & A & $>1$ & $1$ \\ 
\cite{intermittent2009} & SAT & dynamic & A & $>1$ &  $>1$ \\ 
\rowcolor{lightgray}
\makecell{
\cite{hutchinson2019unmanned,fink2010online,radak2017moving} \hspace{1cm}\\ \cite{gradientBased2012,sun2008adaptive,varotto2021MMAR} } & AL & static & A & $1$ & $1$ \\ 
\makecell{\cite{shahidian2017single,hoffmann2009mobile,park2020cooperative}\\ \cite{vander2015algorithms}} & AL & static & A & $1$ & $>1$ \\ 
\rowcolor{lightgray}
\cite{dogancay2012uav} & AL & static & P & $1$ & $>1$ \\
\cite{information_focalLength,realTime_active_vision} & AT & dynamic & P & $1$ & $1$\\ 
\rowcolor{lightgray}
\cite{robotEcology,haubner2019active,varotto2020probabilistic} & AT & dynamic & A & $1$ & $1$\\ 
\cite{koohifar2016receding,koohifar2018autonomous} & AT & dynamic & A & $1$ & $>1$\\ 
\rowcolor{lightgray}
\cite{van2020lavapilot} & AT & dynamic & A & $>1$ & $1$ \\ 
\cite{meera2019obstacle} & PTS & static & P & $>1$ & $1$ \\ 
\rowcolor{lightgray}
\cite{optimal_information_search} & PTS & dynamic & P & $1$ & $1$ \\ 
\cite{bourgault2003coordinated} & PTS & dynamic & P & $1$ & $>1$ \\ 
\hline
\multicolumn{6}{l}{\makecell[l]{\footnotesize SAT: Search and Tracking; AL: Active Localization;\\ AT: Active Tracking. A: Active; P: Passive.\\ 
$N$: number of targets; $M$: number of sensing platforms.
}} \\
\end{tabular}
\end{table}

We say that a target is active when it occasionally~\cite{koohifar2018autonomous} or continuously~\cite{shahidian2017single} releases information on their presence~\cite{cooperative_RinnerCavallaro} (e.g., through radio ID~\cite{canton2017bluetooth}) or on their position (e.g., GPS coordinates~\cite{small_birds}).  
To this aim, target and (robotic) sensing platforms may establish some form of direct communication, for instance via radio~\cite{robotEcology} or acoustic~\cite{wang2018tracking} devices. 
Alternatively, active targets propagate information on their position via radio broadcasting~\cite{fink2010online} or by spreading specific materials in the surroundings (e.g.~hazardous chemicals, fluids, or gaseous particles~\cite{hutchinson2019unmanned}). 
The presence of active targets is usually captured by robots equipped with ad-hoc sensing capabilities (i.e., radio receivers~\cite{van2020lavapilot}, microphone arrays~\cite{haubner2019active}, gas detectors~\cite{park2020cooperative}); once the target is perceived, its position is inferred through the extraction of specific features from the collected data (e.g., the signal strength in radio communication~\cite{li2018indoor})
Active targets often carry electronic devices
that send information to the APE system (e.g., portable WiFi transmitters~\cite{carpin2015uavs}). Despite this, it is important to remark that a target is not required to carry an emitting device to be active: it can be a device-free speaker~\cite{DeepRL_gazeControl} or a radiation source in critical environments~\cite{gao2018robust}. In addition, it should be noticed that the definition of active target does not include any awareness or collaboration guarantee from the target to the sensing platform.

As opposed to active targets, passive ones neither send their location information to robots nor hide from spread substances~\cite{cooperative_RinnerCavallaro,deak2012survey}; thus, they can only passively reflect the electromagnetic waves emitted from external sources (e.g., the platform transmitters~\cite{rosic2020passive}). For this reason, passive targets are usually detected by sensors like cameras~\cite{ryan2010particle} or radars~\cite{white2008radar}, via video processing techniques~\cite{yolo,viola2001rapid} or through the variation of a measured signal (e.g., Differential Air Pressure~\cite{deak2012survey}).

\subsection{The sensing platforms}\label{subsec:platform}

Active sensing can only be accomplished with one or more sensing platforms with self-regulation capabilities~\cite{bajcsy2018revisiting}.
Let $M$ be the number of sensing platforms (robots) whose state is $ \vect{s}_{t,m} \in \mathcal{S}_m$, where $\mathcal{S}_m$ is the state space of the $m$-th robot\footnote{The same considerations made in Sec.~\ref{subsec:targets}, regarding the number of targets, can be repeated for the number of platforms, since $M$ is not always known beforehand and it may change over time.}. 
The state space represents the platform degrees of freedom. It is {\em Euclidean}, $\mathbb{R}^d$, when the platform translates in a one~\cite{noori2016constrained} (i.e.~$d=1$), two~\cite{park2020cooperative} (i.e.~$d=2$) or three~\cite{ramirez2014moving} (i.e.~$d=3$) dimensional space; in this case,
\begin{equation}
    \vect{s}_{t,m} = \vect{c}_{t,m} \in \mathbb{R}^d,
\end{equation}
where $\vect{c}_{t,m}$ denotes the platform position.
The state space is a {\em rotation manifold} when the platform can only rotate, either in one direction~\cite{optimal_information_search} (pan/tilt only platform) or in two~\cite{DeepRL_gazeControl} directions (pan-tilt platform); it follows that
\begin{equation}
    \vect{s}_{t,m} = \begin{bmatrix} \alpha & \beta \end{bmatrix}^\top \in \mathbb{S}^2,
\end{equation}
where $\alpha$ and $\beta$ are the pan and tilt angles, respectively; $\mathbb{S}^2$ is the sphere in $\mathbb{R}^3$.
Finally, the state space is  {\em Riemannian}~\cite{tron2008distributed} when the state embeds both translations and rotations~\cite{MTS}.
There are also high-dimensional robotic systems, like eye-in-hand manipulators~\cite{radmard2017active}, where the state represents the joint parameters vector and encodes a large amount of degrees of freedom.

In APE tasks, the detection event is the process of collecting informative measurements on a specific target~\cite{sanmiguel2017efficient}. To this aim, each sensing platform is endowed with detection capabilities (e.g., object detectors in camera systems~\cite{yolo}, or gas detectors in STE applications\footnote{STE includes estimation of location, emission rate and other variables, needed to describe the spread of the hazardous materials (HAZMAT) using dispersion models.}~\cite{park2020cooperative} - see. Sec.~\ref{sec:applications}). Hence, we can characterize the detection event of target $\ell$ from platform $m$, at time $t$, as a Bernoulli random variable $D_t^{m,\ell} \in \{0,1\}$ with success probability~\cite{varotto2021probabilistic}
\begin{equation}\label{eq:detection_success}
 p(D_t^{m,\ell} = 1 | \vect{p}_{t,\ell},\vect{s}_{t,m}) = P_d(\vect{p}_{t,\ell},\vect{s}_{t,m}).    
\end{equation}
The target detectability depends on both the target position and the sensor state.

\subsection{Active Search and Probabilistic Target Search}

Active Search of target $\ell$ is accomplished before a target detection event has happened yet in any sensor, that is
\begin{equation}\label{eq:search_condition}
    D_k^{m,\ell} = 0; \; k=0,1\dots,t; \; \forall m \in [1,M].
\end{equation}

In other words, Active Search (AS) starts with no target position information available and may employ a deterministic~\cite{mohamed2013coordinated}, random~\cite{cao2006cooperative,gage1994randomized,isler2005randomized} or probabilistic~\cite{optimal_information_search} search. 
Deterministic search follows predetermined patterns, namely a systematic traversal of the whole environment along a predefined direction\footnote{The reader may refer to~\cite{search_pursuit_review,detection_tracking_survey} for a broader discussion on all possible active search classes.}. Examples of deterministic patterns are the spiral search~\cite{fricke2016distributed,bernardini2017combining} and the lawn-mower pattern~\cite{bopardikar2007cooperative,ousingsawat2007modified,hutchinson2019source}.
On the contrary, random search is based on a completely random choice of the next platform actions. 
Randomization 
allows to solve problems that are not solvable by deterministic algorithms~\cite{isler2005randomized}; in spite of this, random search is not always efficient and usually serves as a lower bound when the objective is to minimize the expected time until target detection~\cite{search_pursuit_review}. \textit{Probabilistic Target Search} (PTS)~\cite{search_pursuit_review,furukawa2012autonomous,optimal_information_search} 
accounts for target motion and sensing uncertainties~\cite{optimal_information_search} and formulates the search task within Bayesian probabilistic frameworks (e.g., Recursive Bayesian Estimation - RBE~\cite{smith2013MonteCarlo,PF_tutorial,hue2002tracking}, Bayesian Optimization - BO~\cite{brochu2010tutorial,snoek2012practical,frazier2018tutorial,meera2019obstacle}). In this way, it is possible to encode and keep updated the knowledge about potential target locations as a probability distribution, also referred to as belief or probabilistic map~\cite{liu2015model}; this is done by treating no-detection observations (i.e., measurements with no information on target position) as negative likelihood~\cite{negative_information}. PTS methods consider optimization of the expected value of a search objective~\cite{search_pursuit_review}, such as the probability of detection~\cite{optimal_information_search}, time to detection~\cite{MTS}, information gain~\cite{robotEcology}, or distance to the target~\cite{task_drivenVSinformation_seeking,hasanzade2018rf,chung2011analysis}. Probabilistic approaches are suitable to real-world scenarios, especially when resource consumption (energy and time) is critical~\cite{MTS}; this is due to the use of stochastic target motion models~\cite{li2003survey}, combined with the capability of representing realistic perception uncertainties~\cite{radmard2017active,popovic2019enviromental}. Bayesian-based control has been proven to be faster, more robust, and more effective compared to traditional control methods that rely on direct (or filtered) sensor feedback~\cite{bourne2019coordinated}. In addition, relying on probabilistic strategies is necessary when dealing with large-scale search spaces; indeed, the inherent NP-hard complexity of the searching task~\cite{trummel1986complexity,MTS,MTS_Rinner,popovic2019enviromental,hollinger2008proofs} induces deterministic strategies to be computationally intractable for very large-scale environments~\cite{shubina2010visual}, due to their polynomial~\cite{park2001visibility} or exponential~\cite{hollinger2009efficient,schlotfeldt2019maximum} scalability with the size of the environment. Finally, compared to deterministic path planning, PTS strategies are more capable to adjust the platform behavior in response to new information~\cite{popovic2019enviromental,ghods2020multi}; this higher adaptivity properties are proven to attain more accurate target position estimates and lower search times (i.e., higher efficiency)~\cite{hutchinson2019unmanned,bourne2019coordinated}.  
Thus motivated, in this review we focus on PTS approaches; the reader may refer to~\cite{search_pursuit_review,detection_tracking_survey} for a broader discussion on all possible active search classes.

\subsection{Active Localization and Tracking}

Active Tracking starts when condition \eqref{eq:search_condition} breaks, namely
\begin{equation}
\exists (k,m) \text{ s.t. } D_k^{m,\ell} =1.
\end{equation}

Active Tracking is referred as Active Localization when targets are static, namely
\begin{equation}
    \vect{p}_{t,\ell} = \vect{p}_{\ell}, \; \forall \ell = 1,\dots,N.
\end{equation}

Active Localization (AL) and Active Tracking (AT) use measurements 
provided by single~\cite{haubner2019active} or multiple~\cite{koohifar2018autonomous} on-board sensors to control the platform while estimating the target position. Sensing modalities include vision~\cite{ryan2010particle}, radar~\cite{white2008radar}, audio~\cite{haubner2019active} and radio signals~\cite{shahidian2017single}. Measurements may be aggregated through sensor fusion, which can be uni-modal~\cite{koohifar2018autonomous} or multi-modal~\cite{DeepRL_gazeControl}, cooperative~\cite{park2020cooperative} or non-cooperative~\cite{shahidian2017single}. The sensing platform selects the information to extract from the belief map based on a predefined criterion (e.g.~reaching a goal position~\cite{hasanzade2018rf}, maximizing the information gain~\cite{hutchinson2019unmanned,robotEcology}, maximizing the probability of detection~\cite{liu2015model}) and injects it into the controller, in order to generate the next sensor action.

\subsection{Search and Tracking}
When Active Search precedes Active Tracking, it is referred as Search and Tracking (SAT). 
With multiple targets~\cite{intermittent2009}, search and tracking modes are alternated with respect to the level of uncertainty about the detected targets location. 
In fact, while tracking allows to reduce the uncertainty of a detected target, the search mode is more explorative and gives the possibility to find previously undetected targets. Hence, 
in SAT each robot does not lock its operation in a given mode thus facilitating keeping a balance between search and track operations~\cite{furukawa2006recursive,frew2008target}.

In SAT, the problem is to continuously search for (known or unknown) targets and track already located ones, which are two competing tasks~\cite{cooperative_RinnerCavallaro}. In fact, when the sensing robot tracks already detected targets, it can not search for new ones; on the other hand, when it operates in search mode, some targets are left unobserved and their location uncertainty grows. For this reason, typical SAT solutions~\cite{DeepRL_gazeControl,intermittent2009,frew2008target,how2009increasing} apply mode switching policies leveraging on the number of detected targets~\cite{DeepRL_gazeControl}, the uncertainty level and the accuracy associated to each tracked target~\cite{frew2008target}, and the cost of moving from one mode to the other~\cite{intermittent2009}. 

%

\section{The sensing units}
\label{sec:sensing}
 
\subsection{The observation model}\label{subsec:observation_model}

\begin{figure}[t!]
    \centering
        \includegraphics[width=0.3\textwidth]{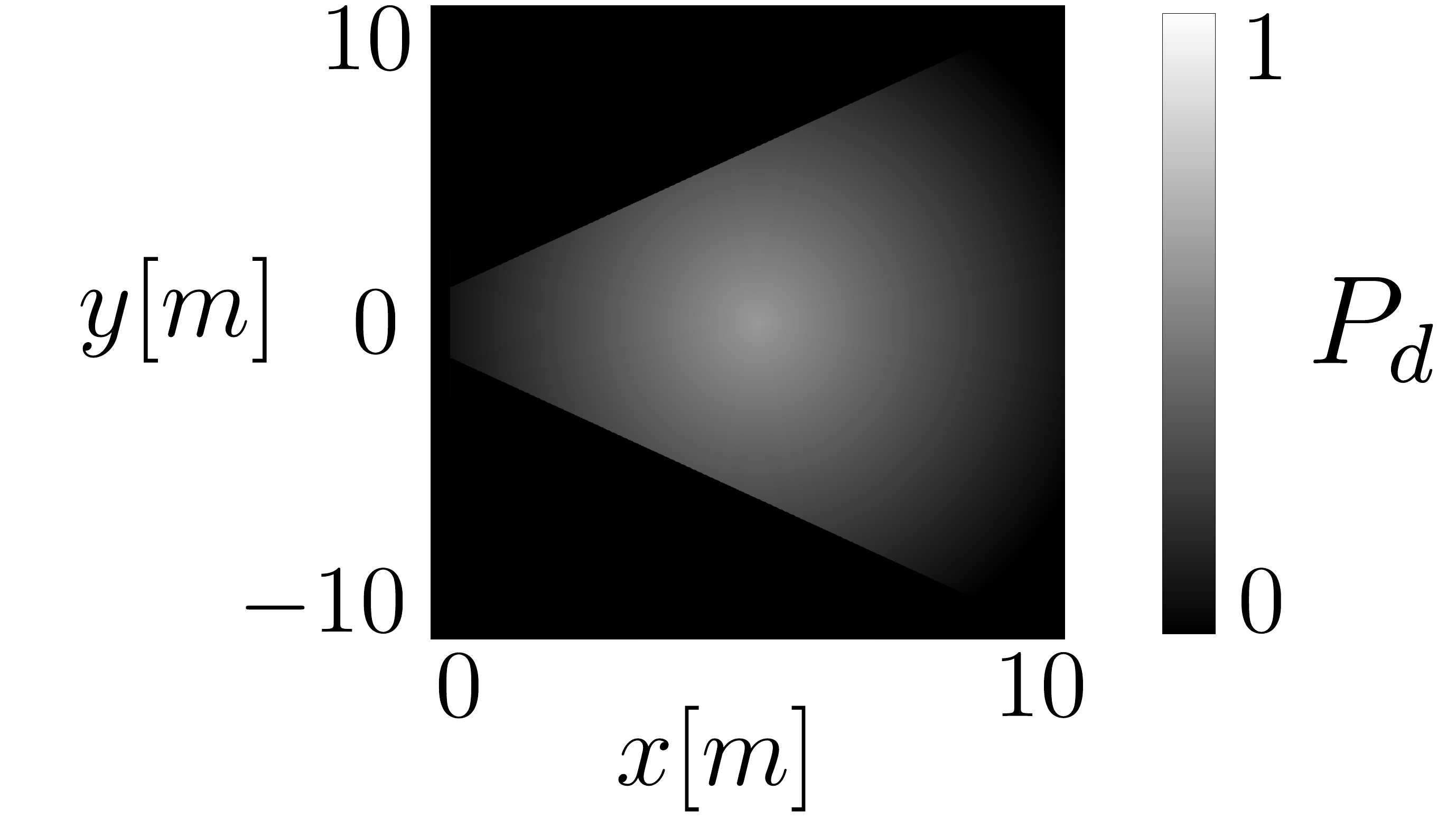} 
        \caption{POD of a visual sensor~\cite{furukawa2012autonomous}: the camera can not detect any object outside the FoV, while the detection capabilities are maximized at the center of the sensing domain. An object is unlikely to be recognized when too far or too close to the camera, due to scale and resolution issues.}
        \label{fig:camera_fov}
\end{figure}

Each platform sensor is characterized by a {\em Field of View} (FoV), $\Phi(\vect{s}_{t,m})$, that is the portion of physical space where the target {\em Probability of Detection} (POD), $P_D$, is positive~\cite{furukawa2012autonomous,sanmiguel2017efficient}, namely
\begin{equation}\label{eq:fov}
    \Phi(\vect{s}_{t,m}) = \left\lbrace \vect{x} \in \mathbb{R}^3 \; | \; 0 < P_D(\vect{x},\vect{s}_{t,m}) \leq 1 \right\rbrace.
\end{equation}
The FoV is a finite solid in $\mathbb{R}^3$, whose shape depends on the POD; this, in turn, is a function of the sensor state and might be non-constant over $\Phi(\vect{s}_{t,m})$. In fact, $\Phi(\vect{s}_{t,m})$ is related to the sensor physical properties, such as the border effects, intrinsic and extrinsic parameters in cameras~\cite{radmard2017active,sanmiguel2017efficient,furukawa2012autonomous}; transmittance in RADARs~\cite{radar_intro}; geometrical configuration of microphone arrays in acoustic systems~\cite{martinson2011optimizing,farmani2015influence,vargas2018impact}; antenna radiation pattern in radio signals~\cite{stoyanova2007evaluation,zanella2016best,stoyanova2007evaluation}. 
The POD is related also to the {\em range} of a sensor, which determines the largest distance at which a target is detectable (over all possible sensor's configurations); formally, for the sensor of the $m$-th platform
\begin{equation}\label{eq:range}
    R_m = \max \left\lbrace \lVert \vect{x} \rVert_2^2 \! : \! \vect{x} \in \mathbb{R}^3; P_D(\vect{x},\vect{s}_{t,m}) > 0, \forall \vect{s}_{t,m} \in \mathcal{S}_m  \right\rbrace
\end{equation}
%
By definition, for the $m$-th platform in position $\vect{c}_{t,m}$ at time $t$, it follows that
\begin{equation}\label{eq:range_POD}
    \lVert \vect{c}_{t,m} - \vect{p}_{t,\ell} \rVert_2^2 > R_m \Rightarrow P_D(\vect{p}_{t,\ell},\vect{s}_{t,m}) = 0,
\end{equation}
namely, a target can not be detected when outside the sensor range.
Fig. \ref{fig:camera_fov} shows the FOV and the POD of a camera~\cite{furukawa2012autonomous}. The limited angle of view restricts the target detectability; border effects reduce the POD along the sides of the FOV, while depth effects make the target less detectable at larger distances (the range is $R \approx \SI{20}{\m}$).
Different sensors are characterized by different FoVs and different ranges \eqref{eq:fov}-\eqref{eq:range}; these, in turn, have a direct impact on the target detection capabilities, according to \eqref{eq:range_POD}.

\vspace{0.1cm}

The behavior of a specific sensor is mathematically characterized through an observation model.
An \textit{observation model} describes the measurements generation process, relates the target state and the sensor configuration with collected observations, and statistically characterizes the noise associated to each measurement. An accurate and reliable observation model is fundamental to apply suitable statistical filtering techniques~\cite{PF_tutorial} that
mitigate sensor nuisance, refine collected data, and quantify the uncertainty associated to an estimation process.
Furthermore, observation models are often at the basis of the controller synthesis in active sensing schemes (see Sec.~\ref{sec:dynamics}). Finally, observation models affect the accuracy, the computational load and the convergence rate of most localization techniques~\cite{PF_tutorial}; therefore, they are crucial and powerful tools for the design of any probabilistic APE solution.
The observation model of sensor $m$ w.r.t. target $\ell$ can be formalized based on the sensor FoV and the target POD  as~\cite{furukawa2012autonomous}
\begin{equation}\label{eq:general_observation_model}
    \vect{z}_{t,m}(\vect{p}_{t,\ell},\vect{s}_{t,m}) = 
    \begin{cases}
        h_m(\vect{p}_{t,\ell},\vect{s}_{t,m},\vect{v}_t), & \text{ if } D_t^{m,\ell}=1 \\
        \emptyset, & \text{ otherwise}
    \end{cases}
\end{equation}
where
$\vect{v}_t$ is the measurement noise, realization of a random vector whose probability density has support in $\mathbb{R}^p$.
When the target is not detected, the platform collects an empty measurement (i.e., $\emptyset$), that is, the observation contains no information on the target. 
Otherwise, a (possibly) non-linear output function 
\begin{equation}\label{eq:measurement_function}
h_m: \Pi \times \mathcal{S}_m \times \mathbb{R}^p \rightarrow \mathbb{R}^p
\end{equation}
transforms the target position into a \mbox{$p$-dimensional} measurement vector. 
The platform can therefore extract information on the target position by leveraging on the set of past and current measurements, namely $\vect{z}_{1:t,m}$.
The analytical form of $h_m(\cdot)$ and the dimension of $\vect{z}_{t,m}$ depend on the specific sensor type; for instance, $\vect{z}_{t,m}$ for a visual sensor is the projection of the target position onto the camera image plane, according to the perspective projection equations~\cite{aghajan2009multi}. 
The output function depends also on the (current) sensor state, since different sensing configurations have different effects on the observation process. An example is given by anisotropic radio receivers, where the communication quality with the target is strongly affected by the relative position and orientation of the sensor with respect to the target~\cite{zafari2019survey}; alternatively, consider the effect of camera intrinsic and extrinsic parameters in visual observations~\cite{aghajan2009multi,sanmiguel2017efficient}. 
The noise components induce a stochastic characterization of the observation model; hence, we can equivalently reformulate~\eqref{eq:general_observation_model} in a probabilistic manner through the \textit{likelihood function}
\begin{equation}\label{eq:likelihood}
\begin{split}
p(\vect{z}_{t,m},D_t^{m,l} \! = \! 1|\vect{p}_{t,\ell},\vect{s}_{t,m})  
& = p(\vect{z}_{t,m}|\vect{p}_{t,\ell},\vect{s}_{t,m})\\
& \times P_D(\vect{p}_{t,\ell},\vect{s}_{t,m}).
\end{split}
\end{equation}
The likelihood represents the probability of gathering the measurement $\vect{z}_{t,m}$, given the platform state and the target position\footnote{For this reason, the observation likelihood is also called \textit{generative model}: each datapoint $\vect{z}_{t,m}$ is a realization of a random vector whose probability distribution function is the likelihood in~\eqref{eq:likelihood}}; in particular, $P_D(\vect{p}_{t,\ell},\vect{s}_{t,m})$ refers to the target detectability, while $p(\vect{z}_{t,m}|\vect{p}_{t,\ell},\vect{s}_{t,m})$ captures the sensor performance, supposed to be independent w.r.t. the target detectability. In case of no-detection observations (i.e., $\vect{z}_{t,m} = \emptyset$), \eqref{eq:likelihood} is referred to as \textit{negative likelihood}~\cite{optimal_information_search,negative_information} and it holds
\begin{equation}\label{eq:negative_likelihood}
    p(\vect{z}_{t,m} = \emptyset, D_t^{m,l} \! = \! 0 \; | \; \vect{p}_{t,\ell},\vect{s}_{t,m}) = 1 - P_D(\vect{p}_{t,\ell},\vect{s}_{t,m}).
\end{equation}
%

\subsection{Probabilistic map}
\label{subsec:map} 

An APE task is said to be probabilistic when information about the target location is modeled with a probability map, kept updated through observations $\vect{z}_{1:t,1:M}$. 
As Fig \ref{fig:APE_scheme} shows, the map is then used to guide the platforms  towards  the  next  actions~\cite{bourgault2003coordinated}.
Probabilistic approaches account for realistic perception uncertainties~\cite{radmard2017active}; hence, they are suitable to manage real-world (noisy) scenarios, unmodeled dynamics, sensing nuisance.
Furthermore, probabilistic decision making has high adaptivity properties~\cite{popovic2019enviromental},
and it is useful when poor a-priori knowledge is available~\cite{ramirez2014moving}
Finally, the stochastic formulation of the measurement process allows to model and account for the lack of data through the negative likelihoods~\eqref{eq:negative_likelihood}.

Most probabilistic APE solutions adopt Bayesian filtering~\cite{chen2003bayesian} or Gaussian Process Regression (GPR)~\cite{rasmussen2003gaussian} for the map generation and update process. In the framework of Bayesian filtering, the map is represented by the posterior distribution of the target location given current and past measurements, namely $p\left(\vect{p}_{t,\ell}|\vect{z}_{1:t,1:M}\right)$. To compute this posterior, RBE is often applied (e.g., Kalman filtering~\cite{shahidian2017single}, particle filtering~\cite{PF_tutorial,smith2013MonteCarlo}). The major benefit of RBE is that it allows to efficiently and recursively update the map, as new measurements are collected. However, the particle-based approximation involved in Sequential Monte Carlo approaches~\cite{smith2013MonteCarlo} makes it difficult to compute uncertainty descriptors (e.g., posterior entropy) from the particle-based approximation of the posterior density~\cite{PF_entropy}. As discussed in Sec. \ref{subsec:criteria}, this is a central issue in APE literature.

Some APE problems are formulated as environmental mapping tasks, where the target location is inferred from the estimated spatial distribution of a physical quantity (e.g., source signal strength~\cite{carpin2015uavs,ghassemi2020extended}) or from occupancy maps~\cite{meera2019obstacle,khan2014information}. In this framework, a common approach is to tesselate the workspace $\Pi$ into a grid map and apply RBE to each cell of the grid~\cite{khan2014information}. 
The main drawback of this method is the poor scalability property in large-scale scenarios or when high spatial resolutions are required.

As an alternative to tessellation or RBE, the target posterior distribution can be estimated directly over the continuous domain $\Pi$ via GPR. Indeed, Gaussian Processes (GP) encode spatial correlation structures in a probabilistic and non-parametric manner; they are also capable of accounting for prior knowledge and data dependencies. Furthermore, GPs provide predictions with uncertainty measures, useful to guide robots towards more uncertain areas.
GPs were initially unpopular for online APE planning applications, due to their computational complexity and non-recursive nature, as well as their bad scaling properties with large datasets. 
Indeed, a critical problem for the long term tasks is the large-scale accumulated data, which might exceed the capacity of onboard computational hardware. 
To alleviate the computational bottleneck caused
by large-scale data accumulation, some 
methods for reducing the computing burdens of
GPs have been recently proposed. For example, sparse GPR dynamically selects only the most informative data, while abandoning the samples that are less
informatively novel~\cite{ma2017informative} (i.e., only a subset of data that provides the greatest contribution). Other approaches suggest efficient and online GP model refinement techniques hinging on Bayesian recursive algorithms~\cite{popovic2019enviromental}. 

\subsection{Perceiving passive targets}\label{subsec:non_cooperative_targets}

Assuming the target to be equipped with ancillary emitting devices is a constraining hypothesis in many real-life applications. For instance, in SAR missions and video surveillance the objective is to detect people (or objects) that, in general, do not establish any kind of communication with the sensing platforms (i.e., the targets are passive). 
Sensors for perceiving passive targets include cameras,  RADARs, and LiDARs.

\vspace{0.1cm}

{\em Vision}-based control, or visual servoing, is widely used for target search and tracking~\cite{ryan2010particle,ramirez2014moving,
radmard2017active,furukawa2012autonomous,
chung2009coordinated,MTS,MTS_Rinner}. The main motivation stems on the affordable cost of cameras and the high information content of visual data (i.e., images); furthermore, various computer vision applications have greatly benefit from the recent advances in signal processing and artificial intelligence (deep learning, in particular), such as object detection and classification, action and activity recognition, and human pose estimation~\cite{voulodimos2018deep}. 
In spite of this, occlusions and FoV directionality limit the range, applicability and success of camera-only platforms~\cite{mavrinac2013modeling}.
Another downside of this technology lies in the fact that adverse weather conditions, e.g., fog or rain, or external light sources, can severely affect its output.

Visual servoing systems use feedback information extracted from  a camera to control the motion of a robot~\cite{radmard2017active}, or the parameters of the camera itself (e.g., pan, tilt, zoom, exposure)~\cite{realTime_active_vision}. 
In this latter case, visual servoing is referred as \textit{active vision}~\cite{bajcsy2018revisiting,chen2011active,bajcsy1988active}.
To avoid inefficient search strategies, characterized by local minima and unstable behaviors~\cite{radmard2017active}, POD modeling is a critical aspect when dealing with vision-based systems~\cite{sanmiguel2017efficient,meera2019obstacle}. The fundamental aspects every realistic camera observation model should consider are image edge effects, detection depth range, and scale effects: the image plane has poor detection capabilities on the edges, due to distortions caused by the camera sensor~\cite{chung2009coordinated}; distant objects are seen at low resolution and their detection probability is low~\cite{furukawa2012autonomous,meera2019obstacle,ghods2020multi}; finally, objects too close to the camera may not be recognized, due to poor scale-invariance properties of some computer vision algorithms.

To account for these phenomena, a possible detection model is~\cite{radmard2017active} 
\begin{equation}\label{eq:detection_model_radmard2017active}
\begin{split}
& P_D(\vect{p}_{t,\ell},\vect{s}_{t,m}) = \Gamma(d_t) \prod_{i=1}^2 \frac{1}{c(u_t,\ell_u) c(v_t,\ell_v)} \\
& c(x,y) = 1+e^{(-1)^i\iota (x - y)} \\
& \ell_u = (-1)^i(L_u/2-\epsilon), \; \ell_v = (-1)^i(L_v/2-\epsilon) \\
& \Gamma(d_t) = \left[ (1+e^{\beta_s (d_t-d_s)})(1+e^{\beta_\ell (d_t - d_\ell)}) \right]^{-1}
\end{split}
\end{equation}
where $L_u$ and $L_v$ represent the image plane dimension; $(u_t,v_t)$ is the target position projected onto the image plane; $d_s$ is the shortest distance at which the target is detectable; $d_t$ is the target distance; $\iota$, $\epsilon$, $\beta_s$ and $\beta_\ell$ are user-defined parameters. Alternatively, the POD can be modeled as a bi-variate Gaussian distribution over the camera image plane, resembling Fig. \ref{fig:camera_fov}~\cite{liu2015model,liu2017model}. The depth effect can be accounted separately by modeling the POD as piecewise continuous function of the sensor altitude~\cite{ramirez2014moving}.
Modeling the POD as in \eqref{eq:detection_model_radmard2017active} or through  Gaussian functions has a strong intuitive rationale and it often works in practical applications. Nonetheless, these models incur in a formal issue: the infinite support that characterizes $c(x,y)$ and Gaussian functions is not compatible with the limited sensing domain of cameras. In this regard, literature offers more realistic camera models~\cite{MTS} that handle the limited domain of visual sensors, but they often require extensive user-defined parameters tuning. 

\textit{LiDAR} and \textit{RADAR} are remote sensing devices used for detection, tracking, and imaging of various objects, especially in agriculture and advanced driving assistance systems (ADAS)~\cite{raj2020survey}. They both use a transmitter to emit a wave (light for LiDAR, radio for RADAR) and, by measuring the time it takes the wave to hit an object and bounce back to the receiver, they can measure the distance to the object. 
Both technologies are highly effective at night, but
RADAR scanners~\cite{optimal_information_search,dogancay2012uav} are cheaper, have longer detection ranges (up to $\SI{200}{\,m}$) and allow for measurements with higher reliability and robustness. In particular, radio waves have low attenuation, which allows them to travel minimally disturbed even in poor weather and without being affected by light conditions.
The main problem of  RADAR technologies is the high sensitivity to environmental disturbances, interferences from external sources, and cluttering, as most radio-based detection systems; moreover, the SNR is strongly related to the target distance and to its scattering coefficient (that, in turn, is related to the material reflectivity)~\cite{radar_intro}. Finally, RADARs can not distinguish different types of objects. 

On the other side, light signals work in the nanometer range, making LiDAR systems much more accurate and precise than RADAR systems. A lower wavelength also means that a LiDAR system can detect smaller objects. Furthermore, high resolution and real-time 3D representation of the environment (point cloud) is helpful to obtain shape and distance of surrounding objects, thus facilitating object detection and classification.
The downside of LiDARs is the strong sensitivity of the measurement process induced by light waves medium. For example, LiDAR systems do not perform as well as RADARs in bad weather conditions.

\subsection{Perceiving active targets}\label{subsec:cooperative_targets}

Active targets propagate in the environment position-related information (e.g., substances, sound, radio frequencies) from which the sensing robots can perceive their presence. 
Most of the works concerned with active targets typically assume the target to be always inside the sensing range; therefore, they neglect the search phase~\cite{koohifar2018autonomous}. Existing works mainly differ in the way target-informative data are used to perform AL and AT. 

\textit{Concentration sensors} such as gas~\cite{park2020cooperative} and atmospheric~\cite{hutchinson2019unmanned} particles detectors are embedded in autonomous sensing vehicles employed in STE tasks. Traditionally, a network of static sensors are used to estimate the source term; however, it is infeasible to cover the entire workspace with sensors dense enough to determine the source before it has spread significantly. Hence, recently mobile unmanned platforms have been equipped for STE. Mobile sensors provide the additional ability to perform boundary tracking of the contaminant and source seeking to aid in the emergency response. Nevertheless, using mobile sensors for STE introduces additional research challenges, concerning how to optimally move each platform in order to produce the best estimate of source parameters within the minimum amount of time, or effort. The solution requires an interdisciplinary fusion of a number of robotics research areas such as autonomous search, multiple robot cooperation, informative path planning, and control.

{\em Acoustic} signal-based localization techniques often leverage on the energy of the incoming signal, or on the time-related information (i.e., time stamps) contained in modulated acoustic signals; from this information, the microphones array mounted on the robot can estimate the transmitter-receiver relative distance or bearing~\cite{cobos2017survey}. MUSIC~\cite{schmidt1986multiple} or SRP-PHAT~\cite{dibiase2000high} algorithms are typically used to estimate the relative Angle of Arrival (AoA) between the robot and the source. Alternatively, AoAs can also be measured sequentially, for instance by a moving robot equipped with microphone arrays~\cite{haubner2019active}. On the other hand, inferring distance information from microphone data is generally difficult in reverberant rooms and often relies on labeled training data~\cite{brendel2018learning}. In general, audio signals enable high localization accuracy~\cite{haubner2019active} and are characterized by wider FoVs with respect to visual ones~\cite{DeepRL_gazeControl}; however, this technology is affected by critical pitfalls, like sound pollution~\cite{wang2019audio}
and extra hardware requirements, which need accurate microphones synchronization, achieved either through intensive pre-deployment phases or with the employment of specialized boards. 
Moreover, acoustic signals are subject to distortions caused by environmental changes, object obstruction, signal diffraction, as well as the ego-noise generated by motors and propellers of the sensing robot itself~\cite{wang2019audio}.

{\em Radio-frequency} (RF) signals~\cite{zafari2019survey,zanella2016best} represent an alternative to acoustics. RF signals guarantee higher reception ranges ($\sim \SI{100}{\m}$)~\cite{zanella2016best}, low energy consumption~\cite{siekkinen2012low} and they often come as parasitic in many real-life scenarios, since most of the current smart phones, laptops and other portable user devices are WiFi or Bluetooth enabled. For these reasons, RF-based AL has been largely used in literature~\cite{shahidian2017single,koohifar2016receding,koohifar2018autonomous,gradientBased2012,fink2010online, intermittent2009} and in commercial applications, especially when the objective is to reduce the deployment, economic and energy impact on the overall setup. In spite of this, latencies~\cite{BLE_performance_review,wifi_latency}
and environmental interference (e.g., cluttering and multi-path distortions) generate strong signal instabilities and limit the accuracy of RF-based solutions~\cite{zafari2019survey,zanella2016best}. To increase the SNR (often below $\SI{10}{\dB}$~\cite{RSSI_SNR}), complex pre-processing algorithms have been suggested~\cite{li2018indoor,zafari2018novel,radak2017moving}.
These, however, are computationally demanding and may need dedicated and energy-harvesting hardware, which collides with the typical use of RF architectures. 
At the same time, latency effects can impose significant challenges to real-time localization; the main cause of latency resides in the adopted communication protocols, which tune collision phenomena and scanning intervals to achieve balance between energy and accuracy~\cite{treurniet2015energy,kindt2017neighbor,rondon2019understanding}. 

Similarly to acoustics, Angle of Arrival (AoA)~\cite{xiong2013arraytrack} can be estimated also between radio transmitters and receivers, using antennae
arrays at the receiver side. To do this, the time difference of arrival (TDOA) at individual
elements of the antennae arrays is computed. To estimate the target (transmitter) position from AoA samples, at least two receivers are required. Then, triangulation or Kalman filter can be applied~\cite{intermittent2009}. The former is a pure geometric approach whose accuracy is limited by the number of sensor platforms and it does not improve with time. On the contrary, Kalman filter is capable of gradually increasing the localization accuracy in time, independently from the number of sensor platforms used~\cite{intermittent2009}. Although AoA can provide accurate estimation when the transmitter-receiver distance is small, it requires complex hardware architectures, as well as decent computation power and energy~\cite{zafari2018novel}. In addition, accuracy deteriorates with increase in the transmitter-receiver distance, where a slight error in the angle of arrival calculation is translated into a large error in the actual location estimation. 

To reduce hardware requirements and the computational load, it is possible to exploit side effects of wireless communication. One of the simplest and widely used techniques is to extract the Received Signal Strength Indicator (RSSI) from standard data packet traffic~\cite{zanella2016best}. The RSSI is the actual signal power strength measured at the receiver side, and, in theory, its value is connected to the distance between transmitter and receiver. More formally, the reference model to describe the relation between the RSSI and the Tx-Rx distance is the so-called \textit{log-distance path-loss model} (PLM)~\cite{goldsmith_2005}, according to which the received power $r(\cdot)$ at distance $d$ from the transmitter can be expressed (in \SI{}{dBm}) as
\begin{equation}\label{eq:PLM}
r(d) = \kappa -10n\log_{10}(d/\delta) + v, \; v \sim \mathcal{N}(0,\sigma^2).
\end{equation}
The zero-mean Gaussian noise $v$ accounts for sensor distortions, model inaccuracies and environmental interference; $n$ is the attenuation gain and $\kappa$ is the RSSI at the reference distance $\delta$. 
Given the RSSI-distance relation \eqref{eq:PLM}, several techniques are applied to estimate target position. \textit{Geometric} approaches leverage on model inversion and trilateration~\cite{wightedTrilateration}; hence, at least three receivers are required. Being deterministic, their robustness to measurement noise and model inaccuracies is often low. Model inaccuracies are particularly critical, since the PLM is too simple to accurately capture the strong signal instabilities and the environmental sensibility~\cite{li2018indoor}.
\textit{Statistical} strategies better account for both model inaccuracies and measurements noise~\cite{coluccia2014rss}, by treating RSSI observations as random variables
and leveraging on statistical filtering techniques to refine raw datapoints (e.g., Kalman filter~\cite{inertialBLELoc}, particle filter~\cite{cenedese2010low,patel2018RSSImap}, belief function theory~\cite{achroufene2018rss}). Model inversion is not always needed and, under specific conditions, one receiver is sufficient to localize a moving target~\cite{varotto2021probabilistic}.
In \textit{Data-driven} models the RSSI-distance relation is represented by a Deep Neural Network (DNN), whose parameters are optimized during the offline training session~\cite{li2018indoor,choi2019unsupervised}. The main benefit comes from the theoretical function approximation properties of DNNs~\cite{hornik1991approximation}: the degrees of freedom given by DNN hyperparameters (e.g., number of layers, number of nodes per layer) allow to tune the level of complexity of the model; hence, in principle, it is possible to accurately capture the underlying propagation behavior. In practice though, hyperparameters optimization is achieved either via extensive brute-force search or through sub-optimal optimization techniques~\cite{wu2019hyperparameter}. Another potential drawback of deep models is that explicit constraints of the underlying physical laws are generally not considered, and this may limit the generalization capabilities~\cite{hu2020physics}. In addition, training processes commonly require larger datasets than model-based parametric estimation procedures. Finally, when input noise is high (i.e., low SNR), overfitting issues may arise.
\textit{Model-free} techniques mainly refer to the fingerprinting strategy~\cite{positionIndoorSurvey,7450196}: the operational space is divided into cells and a database is generated, such that each cell is associated to a specific range of RSSI values. When online RSSI data are collected, matching algorithms are applied to retrieve the target position (at cell resolution). Recently, several works have proven the effectiveness of using Recurrent Neural Networks (RNNs) to guide the fingerprinting system~\cite{hoang2019recurrent,chen2019wifi,sahar2018lstm}. The main disadvantage of model-free techniques is that they may overfit, thus offering poor adaptation capabilities in dynamic and cluttered environments. Furthermore, the database generation process requires cumbersome experimental campaigns, even though distributed procedures may help to tackle this issue~\cite{xu2019LSTMLoc}.

\begin{table}[t!]
\centering
\caption{Comparison of sensing modalities used for APE tasks.}
\label{tab:sensingModalities_classification}
{%
\begin{tabular}{c |ccccc}
\hline
\multirow{2}{*}{\textbf{Ref.}} & \multicolumn{5}{c}{\textbf{Sensing modality}} \\
\cline{2-6} 
& Vision & Radar & Audio & RF & Others   \\
\hline
\rowcolor{lightgray}
\cite{DeepRL_gazeControl} & \cmark & & DP-RTF & &\\
\makecell{
\cite{shahidian2017single,koohifar2016receding}\\
\cite{koohifar2018autonomous,gradientBased2012}\\
\cite{fink2010online,sun2008adaptive}\\ \cite{radak2017moving,van2020lavapilot}} &  & & & RSSI &  \\
\rowcolor{lightgray}
\cite{liu2017model,liu2015model} &  &  &  & & (a) \\
\cite{robotEcology} &  & & & \makecell[l]{
$\!\!
\begin{cases}
    \text{RSSI} \\
    \text{AoA}
\end{cases}
$
}  & \\
\rowcolor{lightgray}
\cite{hutchinson2019unmanned} &  &  &  & & (b) \\
\cite{park2020cooperative} &  &  &  & & (c) \\
\rowcolor{lightgray}
\cite{haubner2019active} & & & AoA & &\\
\cite{intermittent2009} &  &  & & AoA  &\\
\rowcolor{lightgray}
\cite{optimal_information_search} & & \cmark &  &   & \\
\makecell{\cite{information_focalLength,realTime_active_vision}\\ \cite{ramirez2014moving,meera2019obstacle}} & \cmark & &  &   & \\
\rowcolor{lightgray}
\cite{hoffmann2009mobile} &  &  & & \makecell[l]{- RSSI \\- bearing} & 
(d) \\
\cite{varotto2020probabilistic,varotto2021MMAR} & \cmark &  &  & RSSI  & \\
\rowcolor{lightgray}
\cite{vander2015algorithms} & &  &  & bearing  & \\
\cite{dogancay2012uav} & & \makecell[l]{
$
\begin{cases}
    \text{AOA} \\
    \text{TDOA} \\
    \text{Scan-Based}
\end{cases}
$
}  &  & & \\
\rowcolor{lightgray}
\cite{bourgault2003coordinated} & &  RADAR  &  & & \\
\hline
\multicolumn{6}{l}{\makecell[l]{\footnotesize (a): generic sensor with limited sensing domain; (b): atmospheric sensor; \\(c): gas sensor; (d): magnetic field sensing.}} \\
\end{tabular}
}
\end{table}

To sum up, RSSI-based localization approaches are simple and cost efficient~\cite{radak2017moving}, but they suffer from poor localization accuracy~\cite{konings2017rssi}. Cluttered scenarios make the log-distance path-loss model \eqref{eq:PLM} weakly reliable. Furthermore, multipath fading and indoor noise induce severe RSSI fluctuations, so that the noisy component $v$ hides the informative part on the target distance; in fact, in real-world scenarios $\sigma>\SI{3}{dBm}$~\cite{zanella2016best,inertialBLELoc}. Yet, there are many practical solutions that use RSSI to solve APE tasks (see Tab. \ref{tab:sensingModalities_classification}). Moreover, the signal strength can also be used to estimate the relative angle between the receiver and the transmitter (i.e., bearing); however, in this case motorized directional antennas are required~\cite{varotto2021MMAR,vander2015algorithms}.

\subsection{Discussion}

As highlighted by Tab. \ref{tab:sensingModalities_classification}, the same APE task can be solved by using different sensing modalities (e.g., RSSI, bearing and magnetic field~\cite{hoffmann2009mobile}); in general, which sensing modalities to rely on is an application-dependent choice. 
However, only few of the reviewed works provide formal or empirical comparative analysis to motivate their sensors choice, or to investigate the main pitfalls of alternative solutions. 
It follows that literature still misses a comparative analysis on the differences between the various sensors types. In particular, it would be interesting to compare the performance of search and tracking in case of passive and active targets: intuitively, updating the belief map with the negative information provided by passive targets is expected to produce long search times before the target is detected; on the other hand, technologies for active targets (e.g., RF and acoustic signals) are typically sensitive to measurement noise and might struggle in harsh environments. 

Regarding RF-based methods, it is still unclear the impact that some structural properties have on APE tasks; some of these are long sampling times~\cite{zhuang2016smartphone} (usually higher than $\SI{100}{\ms}$~\cite{radak2017moving,canton2017bluetooth}), delays and packet losses~\cite{rondon2019understanding}, cluttering~\cite{stoyanova2007evaluation}, orientation of the antenna~\cite{zanella2016best}, and energy consumption~\cite{treurniet2015energy}. Only few works investigate the influence of these factors on SAT problems~\cite{gu2015energy,stoyanova2007evaluation}. Even in these rare cases, the data are often collected from static targets and the analyses are mainly empirical; hence, the results can not be generalized to different working conditions.

Finally, most of the reviewed solutions rely on a single sensing modality; hence, more research is needed to prove the benefits of multi-modal approaches.

\section{Multi-Sensor APE}
\label{sec:info_fusion}

Single-sensor strategies can sometimes be inadequate:  active targets are perceived at long ranges, but measurement noise represents a critical bottleneck for the localization accuracy; on the other hand, passive targets do not need to use any ancillary equipment, but they are difficult to be detected in highly cluttered environments. For these reasons, some research effort has been devoted to the design of \textit{Multi-Sensor APE} (MS-APE) systems~\cite{park2020cooperative,intermittent2009}, where information is collected from multiple sensing units. 
In MS-APE schemes, the way data is aggregated and manipulated is
a fundamental issue; indeed, an efficient information
fusion technique can reduce the amount of data traffic, filter noisy measurements, and make better predictions and inferences about the target position. 
In the last two decades, several information fusion methods, architectures, and models have been proposed. Consequently, the literature on this theme is extremely vast and can not be classified in a unique way~\cite{smith2006approaches,nakamura2007information,baltruvsaitis2018multimodal,khaleghi2013multisensor}.

In this Section, the works on MS-APE are classified according to the number of platforms, that is, into single~\cite{varotto2021probabilistic} and multi-platform~\cite{koohifar2018autonomous} strategies. These, in turn, are referred to as uni-modal~\cite{radmard2017active} or multi-modal~\cite{DeepRL_gazeControl}, according to the level of sensors heterogeneity involved (see Fig. \ref{fig:MS_APE_classification}).

\tikzset{
block/.style = {draw, fill=white, rectangle, minimum height=3em, minimum width=3em},
block_transp/.style = {rectangle, minimum height=3em, minimum width=3em}}
\begin{figure}[t!]
\center
\resizebox{0.4\textwidth}{!}{%
\begin{tikzpicture}[auto, node distance=3cm,>=latex',scale=0.7, transform shape]
\node [block] (MS_APE) {MS-APE};

\node [block, below of = MS_APE,xshift=3.5cm] (MSSP_APE) { MSSP-APE};
\node [block, below of = MS_APE,xshift=-3.5cm] (MSMP_APE) { MSMP-APE};
\node [block, below of = MS_APE,yshift=-2cm] (uni) { \nlenv{uni-modal \\ $h_m^{(j)}(\cdot)=h(\cdot),\;\forall m,j$}};
\node [block, below of = uni,yshift=1cm] (multi) { \nlenv{ multi-modal \\ $\exists (j,k) \text{ s.t. } h_m^{(j)}(\cdot) \neq h_{m^\prime}^{(k)}(\cdot), $\\$\; m,m^\prime \in [1,M]$}};

\draw [->] (MS_APE.south) --  node{$M=1$} (MSSP_APE.north);
\draw [->] (MS_APE.south) --  node[xshift=-1.2cm,yshift=0.5cm]{$M>1$} (MSMP_APE.north);

\draw [->] (MSSP_APE.south) |- (uni.east);
\draw [->] (MSSP_APE.south) |-  (multi.east);
\draw [->] (MSMP_APE.south) |-  (uni.west);
\draw [->] (MSMP_APE.south) |-  (multi.west);

\begin{scope}[on background layer]
\draw [dashed,fill=lightgray] ($(MSMP_APE.west) + (-0.2,-0.8)$) rectangle ($(MSSP_APE.east) + (0.2,0.8)$);
\draw [dashed,fill=lightgray] ($(multi.west) + (-0.5,-2)$) rectangle ($(multi.east) + (0.5,2.8)$);
\end{scope}

\node [block_transp, below of = MS_APE] (M_box_annotation) { \nlenv{\textit{Number of}\\\textit{ platforms}}};
\node [block_transp, below of = multi,yshift=1.5cm] (modality_annotation) { \nlenv{\textit{Sensors heterogeneity}}};

\end{tikzpicture}
}
\caption{ MS-APE classification according to the number of platforms and the sensors heterogeneity.}
\label{fig:MS_APE_classification}
\end{figure}
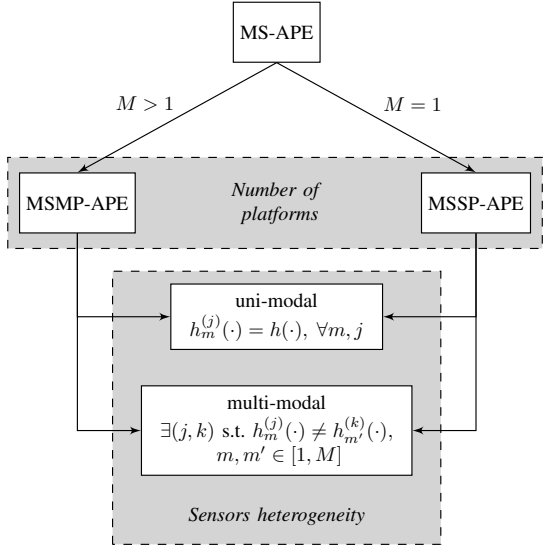

\vspace{0.2cm}

A multi-sensor APE algorithm exploits multiple sensing units to perform search, localization or tracking tasks.   
The general formulation considers $M \geq 1$ sensing robots, with the $m$-th platform endowed with $C_m \geq 1$ sensing units, also referred as channels~\cite{jaimes2007multimodal}. Therefore, a MS-APE system is composed by a total of 

\begin{equation}\label{eq:tot_channels}
    C_{tot} = \sum_{m=1}^M C_m
\end{equation}

sensing units distributed on $M$ spatial locations (i.e., the robot positions). 
In this framework, the measurements collected by every channel are aggregated in a single \textit{multi-sensor observation}
\begin{equation}\label{eq:observation_model_MS_APE}
    \vect{z}_{t}(\vect{p}_{t,\ell},\vect{s}_{t,1:M})  = \begin{bmatrix} \vect{z}_{t,1}(\vect{p}_{t,\ell},\vect{s}_{t,1}) \\ 
    \dots \\ 
    \vect{z}_{t,m}(\vect{p}_{t,\ell},\vect{s}_{t,m})\\
    \dots \\ 
    \vect{z}_{t,M}(\vect{p}_{t,\ell},\vect{s}_{t,M})
    \end{bmatrix},
\end{equation}
where $\vect{z}_{t,m}(\vect{p}_{t,\ell},\vect{s}_{t,m})$ fuses the measurements from the $C_m$ internal channels of the $m$-th platform, namely
\begin{equation}\label{eq:observation_model_MSSP_APE}
    \vect{z}_{t,m}(\vect{p}_{t,\ell},\vect{s}_{t,m})  = \begin{bmatrix} \vect{z}_{t,m}^{(1)}(\vect{p}_{t,\ell},\vect{s}_{t,m}) \\ 
    \dots \\ 
    \vect{z}_{t,m}^{(C_m)}(\vect{p}_{t,\ell},\vect{s}_{t,m})
    \end{bmatrix}.    
\end{equation}
Each channel follows the single-sensor observation model \eqref{eq:general_observation_model}, that is
\begin{equation}\label{eq:general_observation_model_channels}
    \vect{z}_{t,m}^{(j)}(\vect{p}_{t,\ell},\vect{s}_{t,m}) \! = \! 
    \begin{cases}
        h_m^{(j)}(\vect{p}_{t,\ell},\vect{s}_{t,m},\vect{v}_{t,m}^{(j)}),\! & \! \text{ if } D_t^{m^{(j)},\ell} \! = \! 1 \\
        \emptyset, \! & \! \text{ otherwise}
    \end{cases}
\end{equation}
where $\vect{v}_{t,m}^{(j)}$ is the stochastic noise (of given statistics) affecting channel $j$ of platform $m$, while $D_t^{m^{(j)},\ell}$ is the detection event of target $\ell$ from channel $j$ of platform $m$.

\subsection{Uni-modal and multi-modal MS-APE}
\label{subsubsec:MS_APE_uniMulti}

A multi-sensor APE can be either uni-modal~\cite{radmard2017active} or multi-modal~\cite{DeepRL_gazeControl}. A \textit{uni-modal MS-APE} satisfies the condition
\begin{equation}\label{eq:MS_APEuni}
    h_m^{(j)}(\cdot) = h(\cdot), \; j=1,\dots,C_m, \; m=1,\dots,M.
\end{equation}
In this case the system is composed only by homogeneous sensors, namely all platforms channels collect measurements of the same sensing modality. Examples are given by swarms of camera-embedded UAVs~\cite{MTS}, multi-camera eye-in-hand manipulators~\cite{kermorgant2011multi,cuevas2018hybrid}, and multi-antenna platforms~\cite{jais2015review,svevcko2015distance}. 

On the other hand, a \textit{multi-modal MS-APE}  admits the presence of at least two different sensing modalities; 
formally
\begin{equation}\label{eq:MS_APEmulti}
    \exists (j,k) \text{ s.t. } h_m^{(j)}(\cdot) \neq h_{m^\prime}^{(k)}(\cdot), \; m,m^\prime \in [1,M].
\end{equation}
This is the case of a system composed by heterogeneous sensors. Examples are given by robots equipped with cameras surrounded by a microphone array~\cite{DeepRL_gazeControl} (in this case, $m= m^\prime$), or teams of robots where agent $m$ is equipped with different sensors with respect to agent $m^\prime$ (in this case, $m \neq m^\prime$)~\cite{douganccay2009centralized}. 

\subsection{Multi-Sensor Single-Platform (MSSP) APE}
\label{subsubsec:MSSP_APE}

Several APE tasks (e.g., single target tracking~\cite{varotto2021probabilistic}) do not require multiple sensing robots to be accomplished; hence, the use of a single platform (i.e., $M=1$) is more time and cost efficient. 
%
A sensing platform, endowed with multiple sensors on-board, can be either uni-modal or multi-modal. In the former case, it collects the same sensing modality (e.g., visual information~\cite{kermorgant2011multi}) from all its channels, according to the constraint \eqref{eq:MS_APEuni}. Otherwise, if \eqref{eq:MS_APEmulti} holds, it is referred to as multi-modal.
Embedding multiple (potentially different) sources of information on a single platform opens up new perspectives in scene perception. Multi-sensor integration allows to refine single-sensor measurements and obtain more accurate data~\cite{rao2001fusers}. More specifically, uni-modal multi-sensor platforms enable 
parallelisation and specialisation~\cite{esterle2017future}, while sensors heterogeneity induces inherent robustness and complementarity (i.e., different properties of the environment can be perceived). Notably, the aggregated data (either homogeneous or heterogeneous) allow inferences that are not possible with single-sensor measurements~\cite{nakamura2007information}.

An example of uni-modal MSSP-APE is given by multi-camera platforms~\cite{heng2019project}: the multi-view geometry allows stereoscopic vision from single camera observations; this can then be used for localization and 3D geometric mapping tasks~\cite{heng2019project}. At the same time, the possibility of acquiring information from different perspectives allows to better handle occlusions, as well as 
dynamic multi-target scenarios~\cite{hane20173d,cuevas2018hybrid}.
Besides visual-based positioning, uni-modal MSSP-APE schemes are widely used in audio and radio localization as well, as previously discussed in Sec.~\ref{subsec:cooperative_targets}. In this case, an array of receivers (either microphones or antennas) is employed to estimate the target position, while this is not feasible with measurements coming from single receivers only~\cite{xiong2013arraytrack,farmani2015influence}. On top of that, the use of multiple receivers has been proven to guarantee higher accuracy and stability in the localization process~\cite{svevcko2015distance,kleisouris2008impact}. 

In spite of the advantages introduced by uni-modal MSSP-APE solutions, some applications require a high level of complementarity, which can only be obtained through multi-modal platforms~\cite{radio_vision_PF,DeepRL_gazeControl,rajasekaranbayesian}. 
For instance, RSSI observations are often combined with camera measurements~\cite{radio_vision_PF,rajasekaranbayesian} because
visual-based tracking provides high accuracy~\cite{aghajan2009multi} and mitigates the inherent ambiguity in RSSI data~\cite{varotto2020probabilistic}. Conversely, radio observations are used to detect the target at long ranges and under cluttering conditions, providing a rough position estimate and reducing the typical long search phases of camera-only approaches~\cite{ye1999sensor}. Despite being promising for target search and tracking, the fusion of RF and camera sensor data is still an open problem with sparse documentation~\cite{rajasekaranbayesian}.
Finally, it is important to remark that, to achieve multi-modal perception, it is not always necessary to supply the robot with a specific set of physical sensing units, one for each modality; sometimes it is sufficient to extract different features from the same sensor to get different information cues on the environment. For instance, given an array of radio antennas, it is possible to retrieve both ranging (RSSI) and bearing (AoA) data on the target~\cite{dehghan2012aerial}.

The main applications that take advantage from multi-sensor integration systems belong to the field of collaborative robotics and human-robot interaction~\cite{deyle2010rfid,jaimes2007multimodal,DeepRL_gazeControl}. 
Consider, as an example, a robot interacting with surrounding people~\cite{DeepRL_gazeControl}. In this scenario, the robot should be able to decide where to look in order to estimate the number of people in the environment. To this aim, the combination of audio-visual measurements within a deep reinforcement learning (DRL) framework has been proven to enable the robot to autonomously learn gaze control policies~\cite{DeepRL_gazeControl}. In this way, the robot decides where to look next, trying to maximize the number of people present in the camera FoV, and favoring people who speak. 
This application highlights one of the main advantages of audio-visual measurements, that is, the speakers are localized without using any specific device. More specifically, acoustic sources are located via Direct-Path Relative Transfer Function (DP-RTF)~\cite{li2016estimation}, which allows to track people outside the camera FoV and, once a person has been visually detected, it is kept within the visual FoV with high accuracy.
Another application that benefits from mobile sensor-rich robots is autonomous assistance of motor-impaired patients~\cite{deyle2010rfid}.
The robot exploits on-board sensors (RFID tags, cameras, and LiDAR) to localize, approach and grasp objects of interest (e.g., a bottle of water, medications, or a mobile phone), successively delivered to the patient~\cite{calli2018active}. 
%
Finally, multi-modality has been found useful also for STE, with atmospheric sensors providing an initial estimate of the source location, further refined through visual observations~\cite{monroy2018semantic}.

%
%

The stochastic characterization of single channel measurements \eqref{eq:general_observation_model_channels} induces $\vect{z}_{t,m}$ to be a random vector as well. Hence, the techniques used to manage probabilistic maps in single-sensor APE (see Sec. \ref{subsec:map}), can be extended to the multi-sensor (single-platform) case. In particular, once the observation model of each channel is known, it is possible to compute the joint likelihood as~\cite{audio_vision_PF}
\begin{equation}\label{eq:unique_likelihood}
\begin{split}
    p\left(\vect{z}_{t,m}|\vect{p}_{t,\ell},\vect{s}_{t,m}\right) 
    & = p\left(\vect{z}_{t,m}^{(1)},\dots,\vect{z}_{t,m}^{(C_m)}|\vect{p}_{t,\ell},\vect{s}_{t,m}\right) \\
    & = \prod_{j=1}^{C_m} p\left(\vect{z}_{t,m}^{(j)}|\vect{p}_{t,\ell},\vect{s}_{t,m}\right), 
\end{split}
\end{equation}
where measurements coming from different sensors are commonly supposed independent. The joint likelihood is then incorporated into a RBE scheme to estimate the target position.
Notably, this approach allows to treat uni-modal and multi-modal platforms in the same manner.

\subsection{Multi-Sensor Multi-Platform (MSMP) APE}
\label{subsubsec:MSMP_APE}

\begin{table}[t!]
\centering
\caption{Classification of related works on MSMP APE 
}
\label{tab:sensorFusion_classification}
{
\begin{tabular}{c |cc| cc| ccc}
\hline
\multirow{2}{*}{\textbf{Ref.}} & \multicolumn{2}{c|}{\textbf{Data type}} & \multicolumn{2}{c|}{\textbf{Coordination}} & \multicolumn{3}{c}{\textbf{Protocol}} \\
\cline{2-3} \cline{4-5} \cline{6-8}
& U & M &  NC  & C & Di & De  & Ce \\
\hline
\rowcolor{lightgray}
\cite{park2020cooperative,ghods2020multi}
& \cmark &  &  & \cmark &  & \cmark &   \\ 
\cite{intermittent2009,hoffmann2009mobile}
& \cmark &  &  & \cmark & \cmark & &  \\ 
\rowcolor{lightgray}
\cite{shahidian2017single} & \cmark & & \cmark & & & & \cmark \\
\cite{koohifar2016receding,koohifar2018autonomous} & \cmark & & & \cmark & & & \cmark \\
\rowcolor{lightgray}
\cite{vander2015algorithms}
& \cmark &  &  & \cmark  & \cmark & &   \\ 
\cite{dogancay2012uav}
&  & \cmark  &  & \cmark  &  & & \cmark \\ 
\rowcolor{lightgray}
\cite{bourgault2003coordinated}
& \cmark & & \cmark &   &  & \cmark & \\ 
\hline
\multicolumn{8}{l}{\makecell[l]{\footnotesize U: uni-modal; M: multi-modal. NC: non-cooperative; C: cooperative.\\ Di: distributed; De: decentralized; Ce: centralized.}} \\
\end{tabular}}
\end{table} 

In robotics applications, 
the use of multiple sensing platforms (i.e., $M>1$) increases redundancy, making the system less vulnerable to failures and adversarial attacks~\cite{queralta2020collaborative}. Moreover, 
swarms of robots have the ability to simultaneously gather information from disjoint locations, while
the sensing capability of a single robot is restricted to a limited region (see Sec. \ref{sec:sensing}); hence, spatial and temporal coverage problems are reduced when deploying multiple platforms in the monitored environment~\cite{nakamura2007information,chung2018survey}. Finally, multiple platforms are typically employed when the objective includes different concurrent tasks, such as coverage and tracking~\cite{ergodic_exploration}, or SAT~\cite{intermittent2009}. 

In the specific case of APE tasks, the cooperation among multiple agents allows to localize multiple targets, to refine the their position estimate, and to accelerate the convergence rate of the localization process. In addition, by jointly performing decision-making strategies, single-robot motion planning can be optimized to account for other agents actions. To this aim, the sensing robots may share their current observations and exchange information on their state (i.e., coordinated APE)~\cite{cooperative_RinnerCavallaro}. 

As outlined in Tab. \ref{tab:sensorFusion_classification}, MSMP-APE strategies can be classified according to the type of data to be aggregated (i.e., uni-modal~\cite{hoffmann2009mobile} or multi-modal~\cite{douganccay2009centralized}), the type of robot cooperation (i.e., cooperative~\cite{park2020cooperative} or non-cooperative~\cite{shahidian2017single}), and the network communication protocol (i.e., centralized~\cite{koohifar2018autonomous}, decentralized~\cite{park2020cooperative} or distributed~\cite{intermittent2009}).
We will discuss below the main advantages of coordinated APE and provide an overview on the different communication protocols used in multi-sensor settings. 

\subsubsection{Coordination in MSMP-APE}\label{par:cooperative}
Coordination in multi-agent systems implies a minimum level of data exchange\footnote{In this work, non-coordinated MSMP-APE solutions (i.e., those where no communication is established among any couple of agents) are not considered; as a matter of fact, any non-coordinated MSMP-APE architecture can be treated as $M$ independent MSSP-APE sub-systems. }
A coordinated multi-platform APE can be either non-cooperative~\cite{shahidian2017single} or cooperative~\cite{park2020cooperative}. 
\textit{Non-cooperative} 
coordination does not involve any negotiation mechanism and decision-makers plan their actions individually, based on their current knowledge of the world (which is represented by the belief map in probabilistic APE). These kinds of control architectures hide the state of each robot from the other network agents. Therefore, each agent trajectory is computed through independent control laws, implemented internally at each sensing platform~\cite{bourgault2003coordinated}.
The only information exchange allowed is on the observations collected by the agents, namely, the measurements can be spread over the network; in this way, all platforms maintain a synchronized global belief map over time, which serves as common knowledge on which local decisions are made.

\textit{Cooperative} techniques require each agent to take decisions aware of other agents state~\cite{park2020cooperative}, usually, with the aim of achieving a common goal.
Cooperation allows to take the maximum advantage from a multi-agent system; more specifically, cooperative decision making is useful to have balanced environment exploration and resource utilization~\cite{park2020cooperative}, real-time adaptivity levels~\cite{rizk2018decision}, collision avoidance~\cite{shahidian2017single}, as well as fault detection and mitigation capabilities~\cite{yang2019fault}. 
In literature, multi-agent cooperation strategies have been implemented relying on different models and methodological frameworks, including game theory~\cite{game_theory,park2020cooperative}, optimal control~\cite{park2020cooperative},  reinforcement learning~\cite{bucsoniu2010multi}, finite state machines~\cite{intermittent2009},  Markov decision processes (MDP)~\cite{rizk2018decision}, swarm intelligence~\cite{chung2018survey}, and evolutionary computing~\cite{MTS}\footnote{ The reader may refer to~\cite{rizk2018decision} for a comprehensive survey on cooperative multi-agent systems and to~\cite{smith2006approaches,nakamura2007information,khaleghi2013multisensor} for surveys on information fusion in APE-related tasks. }.

\subsubsection{Information \mbox{fusion} protocols}\label{par:protocol}
Multi-platform systems, either cooperative or not, strongly depend on the information fusion technique adopted to spread robots data over the network~\cite{parker2016multiple}. In turn, the functionality of an information fusion strategy greatly depends on robots networking capabilities.
Data aggregation can be achieved through different communication protocols, namely,  centralized~\cite{koohifar2016receding,koohifar2018autonomous,shahidian2017single}, decentralized \cite{lanillos2014multi,bourgault2003coordinated,park2020cooperative}, or distributed~\cite{julian2012distributed,atanasov2015distributed,hollinger2014distributed,intermittent2009}.

{\em Centralized} architectures require the presence of a central base station\footnote{The central base station is capable of communicating with all sensing robots and it embeds sufficient memory and computational resources to store and elaborate the large amount of data received from the sensor network. The base station can be either an external unit, or one of the $M$ sensing robots; in this latter case, the platform needs to be \textit{smart} and with enough on-board capabilities to support the processing, communication and storage demand~\cite{kyung2016theory}.}.
Even though this setup may be practical in specific scenarios~\cite{mukerjee2015practical}, many real-life applications do not allow to rely on a central processing unit, capable of communicating with all network devices and of supporting high computational burdens.    However, a centralized system has a single point of failure (i.e., the base station), which may compromise the operations of the entire architecture. The central node represents also a communication bottleneck: it might not receive complete or updated information, due to sensing and communication limitations, or because of any fault. Finally, centralized protocols suffer from poor modularity and they do not scale well with the network size, since the communication and computation costs increase with the number of agents.
    
{\em Decentralized} architectures are used to increase the scalability and robustness levels with respect to centralized counterparts. Instead of a single node, decentralized protocols convey the information flow on multiple base stations~\cite{ghods2020multi}. In this way, it is also possible to guarantee higher levels of data redundancy within the network; this, in turn, results in improved accuracy, reliability and security. Moreover, the presence of multiple nodes endowed with processing and storage capabilities allows to reduce the computational and memory burden on each single base station. 
    
{\em Distributed} protocols require individual robots to operate with their own measurements and the data received from the communicating nodes. To this aim, each agent of the network can be seen as a base station, with sufficient computational and storage resources on-board.
Distributed architectures are extremely scalable and fault tolerant. Furthermore, they leverage on partially available information; therefore they are only marginally affected by communication problems~\cite{nakamura2007information}. However, distributed optimization does not always guarantee optimality bounds as high as centralized methods do~\cite{yang2019survey,ge2017distributed,tenney1981detection}. The reason is that centralized fusion has a global knowledge (i.e., all measured data is available), whereas distributed fusion generally has only localized information coming from the set of communicating nodes~\cite{nakamura2007information}. 

\subsection{Discussion}

Teams of cooperative~\cite{koohifar2016receding} and (possibly) heterogeneous~\cite{ferrari2009geometric} robots are used to increase the robustness and the efficiency of APE systems~\cite{cooperative_RinnerCavallaro}, through multi-robot cooperation and multi-sensor integration~\cite{park2020cooperative}. Cooperation is particularly useful when the reconstructed belief map has multiple optima; in this case, coordinated motion planning promotes informative joint observations, which enables quick verification and elimination of unlikely target position hypotheses~\cite{bourne2019coordinated}. In spite of this, and although decision making in Multi-Agent Systems (MAS) has seen significant improvements in the past decade, there is only limited research on MS-APE, as highlighted in Tab. \ref{tab:sensorFusion_classification}. This has led to the presence of numerous open research questions that still need to be addressed~\cite{rizk2018decision}, making MS-APE a potential flourishing research field for the next years.
The most interesting research areas 
regard the design of multi-platform schemes that account for on-board limited resources, network scalability, as well as robots heterogeneity. To foster future developments in these sectors, some considerations and suggestions are provided in the following.

\subsubsection{Deployment issues}
The growth of MAS technologies is mainly limited by deployment issues: as a matter of fact, multi-robot systems require more time-consuming deployment procedures, have additional hardware requirements, and come at the cost of higher computational, communication and storage system resources~\cite{hoffmann2009mobile}, especially when cooperative behaviors are required. These factors keep pushing the research and development efforts towards single-platform solutions, when task-compliant.
Nonetheless, a single platform might not guarantee the same effectiveness of a multi-platform system; hence, more research is needed on software and hardware tools to facilitate multi-robot communication and multi-sensor integration through plug-and-play solutions.

\subsubsection{Limited resources}
Robots are commonly characterized by limited on-board resources and they might not be able to offload their computations to the cloud, due to bandwidth scarcity, unreliable connectivity, or minimum latency requirements~\cite{rizk2018decision}. Tightly coordinated tasks also increase the computational burden, due to the large amount of communication and data exchange among agents. 
In addition, most sensing platforms are battery-powered (e.g., UAVs~\cite{meera2019obstacle}); hence, their lifetime is related to the presence of energy-harvesting sensors, the communication activity, and the processors usage~\cite{sanmiguel2016energy}.
Finally, communication and storage limitations might preclude the possibility to collect (or store) updated information from other robots. In conclusion, limited on-board resources reduce the applicability of MAS schemes and make cooperative algorithms challenging in real-life scenarios. Therefore, future research should focus on the development of accurate models that capture the entire spectrum of resource consumption (e.g., energy, bandwidth, storage) as function of the network activity. This models will allow to apply predictive decision-making strategies and guarantee an optimal balance between resource consumption and network operation.

\subsubsection{Scalability}
Another key issue in multi-platform APE algorithms is the system scalability with respect to the number of platforms. In this regard, decentralized and distributed architectures should be considered; notably, they require agents with minimal on-board processing, communication and storage capabilities, but they do not offer the same optimality guarantees with respect to centralized counterparts~\cite{yang2019survey}.
In mobile robotics, distributed optimization is often used to yield agents coordinated motion. In doing so, each sensing robot locally estimates the utility function to be optimized by aggregating its previous observations with those coming from connected agents; the computational complexity is therefore constant with respect to the number of platforms. This approach is however not cooperative, since each agent only considers its own future actions.  
To enable a higher level of coordination, each platform might consider the effect of other robots actions in the objective estimation process. It incurs a linear computational expense in the number of vehicles, yet the effect of the approximation error is provably reduced from that of the single-node approximation, allowing coupled effects between mobile sensors to be captured. Indeed, field experiments have shown better results using the cooperative approach~\cite{hoffmann2009mobile}.
Robots cooperation can be implemented also via finite state machines~\cite{intermittent2009}. A SAT algorithm can start with the team performing a global search and the next platform action might be chosen by optimizing an objective function that accounts for both individual and collective goals. For instance, it may foster group dispersion over the environment, while discouraging fuel consumption, or it should balance the revisiting of the monitored areas. 
Once a target has been detected, the search and track phase starts: robots perform task assignment (to continue the search or to track the detected target), according to the distance to the target, the current number of platforms focused on that target, and the number of targets that have already been detected~\cite{intermittent2009}.

\subsubsection{Heterogeneity}
Robots heterogeneity stems from diverse sensing and actuating capabilities, computing resources, cognitive algorithms, and  morphology~\cite{rizk2018decision,parker2016multiple}. The literature regarding the utilization of heterogeneous robots has shown the benefits of combining either different types of sensors, different perspectives, or different computational and operational capabilities~\cite{queralta2020collaborative,rizk2019cooperative}. More precisely, parallelisation, specialisation and complementarity are important motivations for developing heterogeneity in multi-robot teams~\cite{esterle2017future}. 
For these reasons, heterogeneity is an increasingly prevalent property in cyber-physical and multi-robot systems~\cite{lewis2014s}

When heterogeneity is related to the perception capabilities, it is referred to as multi-modality, as discussed previously in this Section. 
Multiple sensing modalities capture different properties of the environment; thus, the aggregated multi-modal data allow inferences that might be not possible with single-sensor measurements. 
However, multi-modality represents an important challenge to be handled in heterogeneous multi-sensor applications.
As a matter of fact, different data sources require different calibration procedures, each of which might involve extensive human intervention~\cite{zanella2016best,aghajan2009multi}. Recently, some works have proposed self-supervised calibration procedures requiring minimal human effort. These methods hinge on the correlation between the multi-modal data collected by the platform sensors; in particular, one perception channel is used to produce the supervisory signal for the others, enabling the automatic generation of a labeled dataset~\cite{nava2019learning}. 
Calibration is not the only problem to be tackled when dealing with multi-modal systems. Indeed, different data sources suffer from different types and magnitudes of noise, and they might produce unbalanced data. This, in turn, affects the decision making process, which struggles when data have different quality and resolution characteristics~\cite{khaleghi2013multisensor}. Moreover, unbalanced and missing data might affect the belief map generation procedure~\cite{queralta2020collaborative,khaleghi2013multisensor}. 
Multi-modal data sources might also yield conflicting features. For example, visual sensors and RADARs might detect obstacles at different distances~\cite{kumar2020lidar}. 
All these issues are further complicated when multiple robots are involved: the data to be fused are collected in different physical locations and the sensors move with respect to each other. Thus, it is not trivial to evaluate whether different agents are observing the same objects or not, or how to rank observations from different agents~\cite{queralta2020collaborative}.

From the operational perspective, certain applications (e.g., cooperative localization, surveillance, assistive navigation, exploration and mapping) may require the simultaneous use of robots with actuation and morphological heterogeneity~\cite{chen2021collision}. For instance, smaller platforms have access to narrow passages, but their payload is limited~\cite{chen2021collision}; to solve this issue, it is possible to employ larger robots that carry application-specific payloads (e.g., long-duration batteries, heavy sensors or actuators)~\cite{rizk2018decision}.  
Heterogeneous robotic systems represent a fertile research field, where there are still numerous open questions to be addressed. At first, the wide variety of robots being utilized in APE missions, and the different scenarios in which they can be applied, calls for a certain level of collective and situational awareness. This requires the definition of self-organized swarm intelligence models, which account for robots diversity and take into consideration the surrounding environmental conditions. In this framework, the agents of a heterogeneous multi-robot system should identify the main characteristics and constraints of the current environment, and the conditions of the robots they interact with. 
In particular, heterogeneity-aware APE strategies in large scale MAS scenarios, with variable number and agents capabilities, require each robot to automatically learn the capabilities of their peers.
For instance, a UGV collaborating with other UAVs needs to be aware of the different perspectives that UAVs can bring into the scene, but also of their limitations in terms of operational battery lifetime or payload capabilities. 
In general, heterogeneity-aware algorithms, and the explicit incorporation of agents diversity, can be advantageous for multi-robot decision making processes~\cite{davoodi2020heterogeneity}.
Nevertheless, there is still lack of further research in this area, as most existing projects and systems involving heterogeneous robots predefine the way in which they are meant to cooperate~\cite{queralta2020collaborative}.
This is mainly due to the high complexity associated to the design of heterogeneity-aware APE strategies, where robots are meant to autonomously plan their actions, according to the diverse capabilities of their peers. As a consequence, several works introduce additional simplifications or assumptions, such as the hypothesis that the robots sensing and actuation capabilities are a-priori known over the entire network~\cite{chung2009probabilistic}; others design ad-hoc strategies for a MSMP-APE system where the type and the number of robots is predefined~\cite{dogancay2012uav}. 
These assumptions limit the scalability and the fault-tolerance of the system, since they do not account for the possibility of malfunctioning or missing robots~\cite{maza2007multiple}.


\begin{table}[b!]
\centering
\resizebox{0\textwidth}{!}{
\begin{tabular}{c ccc c c}
  &  &  &  &  & \\
\end{tabular}}
\end{table}
\section{Platform dynamics}\label{sec:dynamics}

Active sensing can be defined as an intelligent data acquisition process where a feedback is performed on sensory data (raw or processed) to satisfy specific perception performance measures, such as scene exploration~\cite{ergodic_exploration}, target tracking~\cite{ferrari2009geometric}, and object classification~\cite{active_classification}.
According to~\eqref{eq:general_observation_model}, the measurements are affected by the current platform configuration. This implies that any active sensing scheme must include a decision-making process to dynamically modulate the sensing agent behavior through its state~\cite{bajcsy2018revisiting}.
To this aim, the sensing platform is modeled as a dynamical system in most autonomous perception frameworks~\cite{liu2015model}. In particular, the dynamics of sensing platform $m$, defined by its state $\vect{s}_{t,m}$, is usually described with a deterministic Markovian state transition model~\cite{radmard2017active}:
\begin{equation}\label{eq:sensor_dynamics_general}
\begin{split}
    & \vect{s}_{t+1,m} = q_m\left(\vect{s}_{t,m},\vect{u}_{t,m}\right), \; m =1,\dots,M \\
    & \text{s.t. } \vect{s}_{t,m} \in \mathcal{S}_m, \; \vect{u}_{t,m} \in \mathcal{A}_m
\end{split}
\end{equation}
where $\mathcal{A}_m$ is the {\em control space} (or action set), comprising 
all possible control actions that can be applied by the platform actuators: a finite set of platform displacements in mobile robots~\cite{park2020cooperative}, or a set of Pan-Tilt-Zoom configurations in camera-enabled platforms~\cite{DeepRL_gazeControl,information_focalLength}). The control space represents the platform degrees of freedom (i.e., the controllable state variables), that is, a subset of the state space $\mathcal{S}_m$. The \textit{control input} applied to the platform is denoted as $\vect{u}_{t,m}$; finally, a (possibly) non-linear function  
\begin{equation}
q_m(\cdot): \mathcal{S}_m \times \mathcal{A}_m \rightarrow \mathcal{S}_m
\end{equation}
describes the dynamics of platform $m$ and it mainly depends on the platform structural properties (e.g., aerial, ground or underwater robot)~\cite{radmard2017active,furukawa2012autonomous}. It is worth noting that, in this general setting, each sensing robot has its own control space: as a matter of fact, there may be two agents, say $m$ and $m^\prime$, whose motion capabilities are different (i.e., $\mathcal{A}_m \neq \mathcal{A}_{m^\prime}$). For instance, in heterogeneous camera networks the action set of pan-tilt cameras may be different from that of cameras mounted on aerial robots.   
As introduced in Sec.~\ref{subsec:platform}, the state space can be Euclidean, a rotation manifold, Riemannian or, more in general, a high-dimensional vector. 
In this regard, it is worth noting that, in practice, dealing with a continuous multi-dimensional state space might be computationally demanding and sometimes unfeasible~\cite{ryan2010particle}. For this reason, $\mathcal{S}_m$ and $\mathcal{A}_m$ are usually discretized~\cite{optimal_information_search,realTime_active_vision,van2020lavapilot,chung2009probabilistic}.

Once the platform dynamic model has been defined, it is important to describe the remaining core elements of any active sensing algorithm, namely the control law and the criteria adopted to drive the platform behavior. 

\subsection{The control}\label{subsec:control}

\begin{table}[b!]
\centering
\caption{Classification of related works based on the control space $\mathcal{A}_m$, the set of constraints $\mathcal{C}$ and the optimization method.
}
\label{tab:controlSpace_classification}
{
\begin{tabular}{c| ccc| c| c}
\hline
\multirow{2}{*}{\textbf{$\!$$\!$Ref.$\!$$\!$}} & \multicolumn{3}{c|}{\textbf{Control space}} & \multirow{2}{*}{\textbf{Constraints}} & \multirow{2}{*}{$\!$$\!$$\!$$\!$\textbf{Opt. method}} \\
\cline{2-4}
  & $\!$Pos.$\!$ & $\!$Orient.$\!$ & $\!$Other$\!$ &  & \\
\hline
\rowcolor{lightgray}
\cite{realTime_active_vision} &  & PT & &  & grid-search \\
\cite{DeepRL_gazeControl} &  & PT & & pan speed & Q-learning \\
\rowcolor{lightgray}
\cite{information_focalLength} & & & Z & & grid-search \\
\cite{optimal_information_search} &  & P & & & grid-search\\
\rowcolor{lightgray}
\cite{hoffmann2009mobile} & 2D &  & &  collision avoidance & grid-search \\
\cite{shahidian2017single} & 2D &  & &  \makecell[l]{-collision avoidance\\-connectivity to master} & gradient \\
\rowcolor{lightgray}
\makecell{\cite{gradientBased2012}\\ \cite{sun2008adaptive} \\
\cite{fink2010online}}
& 2D & & & & gradient \\
\cite{van2020lavapilot} & 2D &  & &  \makecell{safe distance from\\ the target} & grid-search \\
\rowcolor{lightgray}
\cite{koohifar2016receding} & 2D & & & turning angle & MPC \\
\cite{koohifar2018autonomous} & 2D & & & & MPC \\
\rowcolor{lightgray}
\cite{liu2017model} & 2D &  & & platform dynamics & MPC\\
\cite{liu2015model} & 2D & & & collision avoidance & MPC \\
\rowcolor{lightgray}
\cite{robotEcology} & 2D & & & & greedy \\
\makecell{\cite{hutchinson2019unmanned}\\ \cite{park2020cooperative}\\ \cite{radak2017moving}}
 & 2D & & & & grid-search \\
\rowcolor{lightgray}
\cite{intermittent2009} & 2D & & & energy consumption & grid-search \\
\cite{haubner2019active} & 2D & & & & MCTS \\
\rowcolor{lightgray}
\cite{varotto2021probabilistic} & 2D & PT & & energy consumption & gradient \\
\cite{vander2015algorithms} & 2D &  & & communication &  heuristic \\
\rowcolor{lightgray}
\cite{dogancay2012uav} & 2D &  & & \makecell[l]{- collision avoidance \\ - agents connectivity   \\ - turn rate    }   & interior point \\
\cite{ramirez2014moving} & 3D &  & &  & grid-search \\
\rowcolor{lightgray}
\cite{meera2019obstacle} & 3D &  & & \makecell[l]{- collision avoidance\\ - time of flight}& BO \\
\hline
\multicolumn{6}{l}{\makecell[l]{\footnotesize P: Pan; PT: Pan-Tilt; Z: Zoom. MCTS: Monte Carlo Tree Search;\\ 
MPC: Model Predictive Control; BO: Bayesian Optimization.
}} \\
\end{tabular}}
\end{table}

In active sensing, the choice of the control input $\vect{u}_{t,m}$ is often cast to an optimization problem of a suitable utility function $J(\cdot)$~\cite{radmard2017active}. This combines the criteria that guide the perception process and it is defined over the control space $\mathcal{S}_m$.
In probabilistic active sensing settings, $J(\cdot)$ depends on some information extracted from the belief map (see Sec. \ref{subsec:criteria}). Sometimes, if the platform is multi-modal and collects observations from $K<C_m$ different sensing modalities, $J(\cdot)$ is computed as the aggregation of $K$ utilities, one for each sensing modality~\cite{dogancay2012uav}. 

Additional constraints may be introduced in the optimization problem in order to account for finite resources availability (e.g., time, energy)~\cite{MTS,nguyen2013energy,noori2016constrained}, communication limitations~\cite{douganccay2009centralized}, or collision avoidance requirements~\cite{MTS_Rinner}. 
The control input is therefore computed as~\cite{hutchinson2019unmanned}
\begin{equation}\label{eq:input_optimization}
\vect{u}_{t,m} = \vect{u}_{t,m}^* = \argmax_{\vect{u} \in \; \mathcal{A}_m \setminus \mathcal{C}_{t,m}}  J\left(\vect{u} \; | \;  \vect{z}_{1:t,m},\vect{s}_{t,m}\right) ,
\end{equation}
where 
\eqref{eq:input_optimization} is computed for each platform, that is for $m =1,\dots,M$; 
the notation $J\left(\vect{u} \; | \;  \vect{z}_{1:t,m},\vect{s}_{t,m}\right)$ says that $J(\cdot)$ is function of $\vect{u}$, given the actual platform state and the sequence of observations until time $t$, namely \mbox{$\vect{z}_{1:t,m} = \{ \vect{z}_{1,m},\dots,\vect{z}_{t,m} \}$}; $(\cdot)^*$ denotes the solution of an optimization problem; $\mathcal{C}_{t,m}$ is the set of constraints for the $m$-th robot at time $t$. Obviously, $\mathcal{C}_{t,m}$ is time and platform-dependent; for instance, if a UGV is close to the borders of a room and another one is at the centre of the same room, the former has constrained motion capabilities, while the latter is completely free to move. 

In \eqref{eq:input_optimization} 
each robot exploits its local information (i.e., $\vect{z}_{1:t,m}$) and behaves independently to the others, since $J(\cdot)$ depends only on $\vect{s}_{t,m}$ (i.e., non-coordinated APE). If a coordinated MSMP-APE is employed,
the objective function depends on all measurements collected by other agents (or a subset of them, if the information sharing protocol is not centralized), according to \eqref{eq:observation_model_MS_APE}. The problem \eqref{eq:input_optimization} becomes

\begin{equation}\label{eq:input_optimization_coordinated}
\vect{u}_{t,m}^* = \argmax_{\vect{u} \in \; \mathcal{A}_m \setminus \mathcal{C}_{t,m} }   
J\left(\vect{u} \; | \;  \vect{z}_{1:t,1:M},\vect{s}_{t,m}\right).
\end{equation}

Finally, if the MSMP-APE is cooperative~\cite{park2020cooperative}, the decision making process is aware of other agents state and actions, namely 
\begin{equation}\label{eq:input_optimization_cooperative}
\begin{split}
& \vect{u}_{t,1:M}^* = \argmax_{\vect{u}_{1:M}}   
J\left(\vect{u}_{1:M} \; | \;  \vect{z}_{1:t,1:M},\vect{s}_{t,1:M}\right), \\
& \text{s.t. } \vect{u}_{1:M} = \{ \vect{u}_1, \dots, \vect{u}_M \}, \quad \vect{u}_m \in \mathcal{A}_m \setminus \mathcal{C}_{t,m}
\end{split}
\end{equation}

where \mbox{$\vect{s}_{t,1:M}= \{\vect{s}_{t,1},\dots,\vect{s}_{t,M} \}$} and similarly for $\vect{u}^*_{t,1:M}$.

\subsection{Control space, constraints and optimization}
\label{subsec:control_space} 

This Section classifies the APE-related literature according to the control space, the principal platform constraints, and the optimization methods adopted to solve the control problem~\eqref{eq:input_optimization}\footnote{The focus is on the basic optimization problem~\eqref{eq:input_optimization}; the variation~\eqref{eq:input_optimization_coordinated} involves a larger amount of available information, while~\eqref{eq:input_optimization_cooperative} extends the problem dimensionality.}; the objective is to highlight how the same APE task can be tackled by leveraging on platforms with different degrees of freedom and by applying different control techniques. 
In fact, as Tab. \ref{tab:controlSpace_classification} suggests, some works exploit approximate~\cite{radak2017moving}, greedy~\cite{robotEcology}, or heuristic control laws~\cite{MTS_Rinner}; others adopt optimal and predictive strategies~\cite{shahidian2017single}, and a few propose statistical black-box optimization methods~\cite{meera2019obstacle}. This variability comes from the fact that the performance metrics and the optimization strategies vary when operating on different action spaces (e.g., rotation manifolds or canonical Euclidean spaces)~\cite{varotto2019distributed}; furthermore, the choice of the control algorithms depends also on the constraints that characterize the platform movements~\cite{MTS_Rinner}. Finally, the cardinality of $\mathcal{A}_m$ has a direct impact on the computational demand and on the optimality guarantees of a control strategy; in particular, small action sets~\cite{van2020lavapilot} and approximation schemes~\cite{realTime_active_vision} are sometimes necessary to meet real-time performance requirements. 

\subsubsection{Platform position}
In many cases it is possible to regulate the planar~\cite{koohifar2016receding} or the 3D~\cite{ramirez2014moving} position of ground~\cite{gradientBased2012} and aerial~\cite{koohifar2016receding,chung2018survey} robots, without any control on the sensors orientation. Some examples are given by fixed radio antennas mounted on UGVs~\cite{gradientBased2012}, atmospheric sensors~\cite{park2020cooperative}, or fixed cameras embedded in UAVs (and usually facing downwards)~\cite{ramirez2014moving}.  
When the cardinality of the action space is small, the optimal control input $\vect{u}_{t,m}^*$ can be chosen according to an extensive search over all possible candidate actions (i.e., {\em grid-search optimization})~\cite{hutchinson2019unmanned}. Grid-search becomes impracticable with continuous action spaces, while {\em gradient-based controllers} are more suitable~\cite{shahidian2017single}. The main limitation of gradient-based algorithms is the sensibility to local optima; hence, their applicability is restricted to convex functions over the control space. However, the presence of hard constraints might violate this condition; furthermore, the need to analytically evaluate the gradient of the cost function represents a further limitation of gradient-based controllers.  
To reduce the sensibility to local optima, several strategies can be adopted. In~\cite{radak2017moving} an UAV is meant to fly a predefined distance towards the peak of the target belief map and, in case of multiple peaks, the heading angle is chosen randomly; the introduction of a stochastic term in the decision process might allow the platform to escape from local optima, at the cost of possible unstable and inefficient trajectories. 
On the other hand, when the gradient of the cost function can not be analytically computed, or it is computationally expensive to evaluate, it can be estimated through the collected data $\vect{z}_{1:t}$ via {\em gradient-assisted} path-planning algorithms~\cite{gradientBased2012,sun2008adaptive}. 
This kind of strategies are completely model-free (i.e., the observation model is not needed), which makes them highly adaptive. However, they also have many limiting factors: first, they are not effective when the underlying data generative model does not follow a convex function; secondly, they can be applied only when the target is static; finally, in turbulent, cluttered environments, or when sensors are noisy (e.g. in RSSI-based source localization~\cite{sun2008adaptive}), the gradient does not always lead directly to the source, with consequent unstable platform behaviors. 
When the form of the objective function is complex and intractable to analyze, non-convex, nonlinear, high-dimensional and noisy, {\em heuristic strategies} can be adopted (e.g., ant colony optimization~\cite{MTS_Rinner}, genetic algorithm~\cite{MTS}). Interestingly, heuristic optimization techniques are particularly suitable to handle platform constraints (e.g., communication and collision avoidance)~\cite{MTS_Rinner}, even though they might produce sub-optimal solutions in terms of tracking performance (e.g., localization error) w.r.t. optimal control approaches~\cite{koohifar2016receding}.

\subsubsection{Platform orientation}
When the platform is static (i.e., its position can not be regulated), the APE task can only be accomplished through rotation movements. This is the case of orientable RADAR scanners~\cite{optimal_information_search} and pan-tilt cameras, which span the surrounding environment 
according to discretized action spaces, composed by a fixed number of angular sectors. 
When the objective is to perform active tracking, but the target dynamics is unknown, Active Disturbance Rejection Controllers (ADRC)~\cite{chen2014novel} can be used: the uncertainties on the target dynamics are estimated and compensated through an extended state observer (ESO), and then treated as system disturbances. As an alternative to ADRCs, Model Predictive Controllers (MPCs) can be used~\cite{saragih2019visual}; MPCs generate optimal control signals, have predictive capabilities, embed input saturation constraints in the objective function~\cite{liu2017model}, and allow to solve non-linear control problems. 
The optimization problem~\eqref{eq:input_optimization} accounts for a one-step look-ahead action; hence, it is referred to as \textit{myopic} path-planning~\cite{douganccay2009centralized}. In \textit{receding horizon} path-planning (i.e., MPC)~\cite{koohifar2016receding,koohifar2018autonomous,liu2017model,liu2015model} the problem defined in~\eqref{eq:input_optimization} is extended to a multiple-step look-ahead horizon.
It is important to remark that MPC usually implies high computational costs, it needs an accurate dynamic model of the target, and model uncertainties are very difficult to resolve\footnote{The reader may refer to~\cite{wang2009model,bemporad2005model} for a comprehensive overview on predictive control}.


\subsubsection{High-dimensional control spaces}
High-dimensional control spaces are common in many robotic systems, like visual servoing architectures and eye-in-hand manipulators~\cite{radmard2017active}. Robots with a large number of degrees of freedom are more robust and more efficient than platforms with limited action spaces. As an example, when the purpose is to visually detect a target, acting on the entire camera pose (i.e., position and pan-tilt orientation) is beneficial~\cite{varotto2021probabilistic}: the camera orientation allows to observe the target even when it is not close to the robot; to maximize the target detectability, it is possible to control the robot position, under the assumption that the POD is inversely proportional to the target-platform distance, as in~\eqref{eq:detection_model_radmard2017active}; the same formula suggests that, if camera sensors are employed, it is also possible to increase the target detectability by acting on the focal length~\cite{MTS,shubina2010visual}. 
The main issue of dealing with high-dimensional control spaces is that gradient-based solvers are typically not effective. This is due to the fact that the objective functions are rarely convex with respect to the control space. Hence, heuristic~\cite{MTS} or grid-search~\cite{shubina2010visual} optimization strategies are commonly applied to solve~\eqref{eq:input_optimization}. 
Usually, to account for joint limits and dynamics constraints, a multiplicative penalty term is introduced to the cost function, or implicitly embedded into the sensor motion model~\cite{radmard2017active}.

\subsubsection{Constraints}
Most existing strategies for APE do not include any constraint in the task formulation~\cite{hutchinson2019unmanned,park2020cooperative,intermittent2009,radak2017moving}, that is, $\mathcal{C}_m =\emptyset$. 
However, in real-life applications the platform movements are limited by dynamics saturation levels~\cite{liu2017model} (e.g., robot speed), while the network communication capabilities are commonly affected by distance-based inter-agent constraints~\cite{MTS_Rinner, vander2015algorithms}. In addition, time and energy resources play a critical role in APE tasks, usually in the form of (soft or hard) specifications to be satisfied (e.g., in UAV-based SAR  missions)~\cite{matthiesen2010efficient,nguyen2013energy,noori2016constrained}. Finally, when multiple platforms are employed, or when the environment is cluttered, the risk of collision among robots~\cite{hoffmann2009mobile}, or with surrounding obstacles~\cite{liu2015model,meera2019obstacle}, should be minimized. 

If the constraints can not be violated at any moment during the APE mission (i.e., hard constraints), they must be explicitly considered in the optimization problem~\eqref{eq:input_optimization}. This is the case of sensing robots with a fixed amount of internal resources, such as battery powered cameras~\cite{varotto2021probabilistic}, or unmanned vehicles~\cite{varotto2021probabilistic}. 
An alternative is to incorporate the constraints into the cost function (i.e., soft constraints), obtaining a new unconstrained optimization problem over a functional that aggregates both APE requirements and constraints 
~\cite{intermittent2009,popovic2019enviromental,meera2019obstacle,vander2015algorithms}. This is particularly useful when robots are allowed to temporarily break their constraints to take better measurements (i.e., move closer to the target, or acquire more informative data).

As suggested by Tab. \ref{tab:controlSpace_classification}, the APE literature offers a wide variety of constrained scenarios; nonetheless, only few of the reviewed works consider energy consumption~\cite{varotto2021probabilistic}, or time of search~\cite{miki2018multi}, as hard specifications. Thus, future research should focus on the impact of hard resource constraints in APE performance~\cite{varotto2021probabilistic}. To this aim, high-dimensional control spaces would be useful for balancing resource consumption and APE task completion~\cite{MTS}. For instance, 
\eqref{eq:detection_model_radmard2017active} suggests that, in camera-based APE tasks, 
the only way to have effective detections is either to zoom-in, or to move closer to the target; however, platform movements are much more energy consuming than the regulation of the camera focal length. Thus, zoom control might have a beneficial impact when dealing with energy-aware~\cite{sanmiguel2014cost,sanmiguel2016energy} and Quality of Sight-aware~\cite{sanmiguel2017efficient} problems, where both detection capabilities and energy preservation matter.  

\subsection{Discussion}\label{subsec:discussion_controlSpace}

As discussed in Sec. \ref{subsec:control},
APE is the control of the perception process, which can be formulated as a mathematical optimization program where the cost function depends on the platform state $\vect{s}_t$. All the articles reviewed in this work solve the control problem~\eqref{eq:input_optimization} assuming perfect knowledge on the platform state. Nonetheless, this is an ideal assumption, since the platform state is often measured from noisy on-board sensors (e.g., IMU units), or it needs to be estimated from other agents~\cite{varotto2019distributed}; in this case, the measurement or estimation inaccuracies might have a significant impact on the active sensing performance~\cite{popovic2019enviromental}. Therefore, future research should focus on the formulation of APE solutions that mathematically couple both target and platforms uncertainties in a single objective function. 


Reinforcement learning (RL)~\cite{hernandez2019survey} is based on a framework that considers an agent that interacts with the surrounding environment, it observes the consequence of any action taken, and it is capable to measure the success of its actions via a reward (/punishment) system.
Experience is used to learn a policy that, given the current state, chooses the next best action to maximize the expected future reward. 
RL is used for  active sensing~\cite{queralta2020collaborative}, whose essence is understanding and adapting to changes in the environment to plan future sensing actions. Interestingly, the mathematical formulation of RL perfectly fits with the APE characterization given in \eqref{eq:input_optimization}: the cost function $J(\cdot)$ represents the expected reward, while $\vect{u}_{t,m}^*$ is the best action to take.
For instance, in camera-based active tracking the reward is typically based on a predefined viewpoint of the target (e.g., a desired viewing angle or a given platform-target distance~\cite{luo2018end}); in some cases the reward considers the uncertainty associated to the target location estimate~\cite{calli2018active}. 
In AS instead, the reward may account for the time of search, the presence of cluttering in the environment, and the probability of detecting the target, which is usually related to the platform-target distance, as in \eqref{eq:detection_model_radmard2017active}~\cite{sandino2020autonomous,meera2019obstacle}.
The main advantage of RL is that the learning process does not require a dataset of labeled input-output pairs; thus, the robot learns without any human supervision. Usually, the robot starts by performing some random actions and gradually learns to follow a desired SAT behavior. 
Another advantage of RL is that it provides an end-to-end approach from sensing to actuation; therefore, it integrates the perception and control aspects within a single framework~\cite{queralta2020collaborative}.
Finally, the RL framework can also be extended to multi-agent and cooperative scenarios (i.e., multi-agent reinforcement learning, MARL~\cite{hernandez2019survey,bucsoniu2010multi}), with direct application to distributed and coordinated MSMP-APE problems. 
Despite the aforementioned benefits, a critical question of reinforcement learning methods is that they require long training sessions, which are not always desiderable in robotics applications~\cite{DeepRL_gazeControl}. Furthermore, the training process of a RL model involves random exploratory actions, which can be potentially unsafe (especially in human-robot cooperation) and costly (in terms of hardware damage)~\cite{arndt2020meta}. These issues are particularly critical in SAR operations, where it is difficult to collect or to create sufficient training data and experiences~\cite{queralta2020collaborative}.
More in general, the dependence on training data may limit the applicability of RL-based solutions in unstructured environments, where very few assumptions can be made on the data generation and acquisition process; hence, the robustness performance obtained during the training phase can not be guaranteed during system deployment. For example, dramatic lighting changing conditions significantly affect the performance of an image recognition algorithm~\cite{calli2018active}; thus, an unstable system is obtained if during training phase it has not explored and experienced all possible lighting conditions. Similar issues may arise in camera-based SAT tasks where the target is unknown or its appearance does not match with the training dataset~\cite{calli2018active}. 
A promising approach to deal with dynamic and unstructured environments is transfer learning~\cite{arndt2020meta,DeepRL_gazeControl,luo2018end}: ad-hoc simulators are employed to train an inference model, which is then fine-tuned and adapted to the real-world environment, since a simulation may not capture all physical phenomena~\cite{arndt2020meta}. As an alternative, it is possible to rely on model-free control strategies (e.g., extremum seeking control~\cite{calli2018active}), which do not utilize training data and implement continuous adaptation strategies to respond to the current state of the environment.
In view of the aforementioned challenges, more research is needed on efficient and robust control techniques in unstructured and cluttered environments~\cite{calli2018active,sandino2020autonomous,meera2019obstacle}.

\section{APE optimization criteria}
\label{subsec:criteria}

In active sensing applied to target search and localization tasks, the platform state must be regulated in order to satisfy several quality of sensing and quality of localization requirements. For instance, a Pan-Tilt-Zoom camera can be controlled to guarantee minimal detection capabilities or maximum probability of detection; in the former case it is sufficient to move the camera so that the target is inside the FoV, namely $\vect{p}_{t} \in \Phi(\vect{s}_{t,m})$\footnote{In this Section the formulas are related to single-platform single-target solutions; hence, to ease the notation, the subscripts used to denote the $m$-th platform and the $\ell$-th target can be omitted.}; in the latter case, it is necessary to model and optimize the POD function through the platform movements. Another example comes from radio-based tracking systems; these often necessitate of automatic algorithms for antennas alignment to optimize the quality of communication between the platform and an active target with which communication has been established~\cite{li2019design}.

In this Section, we classify APE algorithms as information-seeking or task-driven, according to the cost function involved in the optimization problem \eqref{eq:input_optimization} (see Tab. \ref{tab:criteria_classification}).  

\begin{table}[b!]
\centering
\caption{Classification of related works based on the APE criteria.}
\label{tab:criteria_classification}
{
\begin{tabular}{l| c| c}
\cline{1-2}
\textbf{Ref.} & \textbf{APE criterion} &  \\
\hline
\cellcolor{lightgray}
\cite{DeepRL_gazeControl} &  \cellcolor{lightgray}{\makecell[l]{-source detection\\
-people in FoV} }&
\parbox[t]{2mm}{\multirow{10}{*}{\rotatebox[origin=c]{+90}{task-driven}}}  \\
\makecell[l]{
\cite{fink2010online,intermittent2009,radak2017moving,ramirez2014moving} \\\cite{varotto2021probabilistic,van2020lavapilot,liu2017model}
}&  distance to target estimate  \\ 
\rowcolor{lightgray}
\cite{gradientBased2012,sun2008adaptive}  &  highest RSSI value \\
\cite{optimal_information_search}  & expected POD    \\
\rowcolor{lightgray}
\cite{bourgault2003coordinated}  & cumulative POD    \\
\cite{liu2015model}  & \makecell{-distance to target estimate\\-POD}    \\
\rowcolor{lightgray}
\cite{vander2015algorithms}  & mission time \\
\cite{meera2019obstacle} & target detectability\\

\hline

\cellcolor{lightgray}
\cite{shahidian2017single} &  \cellcolor{lightgray}
D/A-optimality & \parbox[t]{2mm}{\multirow{10}{*}{\rotatebox[origin=c]{+90}{information-seeking}}} \\
\cite{koohifar2016receding,koohifar2018autonomous,dogancay2012uav}  &  D-optimality  \\
\rowcolor{lightgray}
\cite{vander2015algorithms} & FIM spectrum\\
\cite{robotEcology,hoffmann2009mobile}  &  MI \\
\rowcolor{lightgray}
\cite{hutchinson2019unmanned,optimal_information_search} &  expected $D_{KL}$  \\
\cite{park2020cooperative} &  infotaxis   \\
\rowcolor{lightgray}
\cite{haubner2019active,information_focalLength,realTime_active_vision}  &   expected $H$   \\
\cite{fink2010online,meera2019obstacle} & predictive variance\\
\rowcolor{lightgray}
\cite{liu2017model} & covariance trace \\
\cite{ramirez2014moving} & conditional $H$\\
\hline
\multicolumn{3}{l}{\makecell[l]{\footnotesize FIM: Fisher Information Matrix; MI: Mutual Information; $H$: Entropy;\\ $D_{KL}$: Kullback-Liebler divergence; POD: Probability of Detection.
}} \\
\end{tabular}}
\end{table}

\subsection{Information-seeking control}\label{subsubsec:information_seeking}

Information-seeking control aims at choosing the next sensor state such that the expected future measurement optimizes a certain information gathering measure; for this reason, the APE strategies belonging to this control class are biased towards an \textit{explorative}  accomplishment of the positioning task\footnote{The reader may refer to~\cite{gupta2006interplay} and \cite{xu2014exploration} for an overview on the concepts of exploration and exploitation.}. 
The most intuitive information-seeking criterion is the minimization of target position \textit{estimate variance}~\cite{chung2009coordinated}. 

Other criteria are based on the optimization of real-valued functions of the \textit{Fisher Information Matrix} (FIM), which is related to the estimation uncertainty ellipsoid~\cite{ucinski2004optimal,vander2015algorithms}. For instance, A-optimality minimizes the trace of the FIM and represents the average variance of the estimate~\cite{shahidian2017single}; D-optimality instead minimizes the log-determinant of the FIM and it is related to the volume of the uncertainty ellipsoid~\cite{dogancay2012uav}. D-optimality is proven to outperform A-optimality in terms of trace of the error covariance matrix~\cite{shahidian2017single}. Both criteria are suitable for gradient-based controllers~\cite{shahidian2017single} and MPCs~\cite{koohifar2016receding,koohifar2018autonomous}. In this latter case, the FIM can be approximated via Monte Carlo numerical methods.
In general, the main issue of FIM-based approaches is that they rely on state estimators, like the Kalman Filter and the Extended Kalman Filter, that are ill-suited when target state and sensor models follow severe non-linear/non-Gaussian distributions. In this context, several studies have proven that particle filters~\cite{gordon2004beyond,koohifar2018autonomous} appear to be more appropriate, reliable and accurate than conventional Kalman-based methods in many tracking applications. 

The information gain obtained with action $\vect{u}_t$ can be also quantified by the \textit{mutual information} (MI) between the target position and the sensor observation, namely~\cite{hoffmann2009mobile}
\begin{equation}\label{eq:MI_hoffmann2009mobile}
\begin{split}
    J( \vect{u}_t | \vect{z}_{1:t} ,\vect{s}_t) 
    & = I\left(\vect{z}_{t+1}(\vect{s}_{t+1});\vect{p}_{t+1}\right) \\
    & = H(\vect{p}_{t+1}) - H\left(\vect{p}_{t+1}|\vect{z}_{t+1}(\vect{s}_{t+1})\right).
\end{split}
\end{equation}
$H(\vect{p}_{t+1})$ is the entropy of the random vector $\vect{p}_{t+1}$, while $H\left(\vect{p}_{t+1}|\vect{z}_{t+1}(\vect{s}_{t+1})\right)$ represents the conditional entropy, that is the expected entropy of the posterior belief, computed as
\begin{equation}\label{eq:conditional_entropy}
\begin{split}
    & H\left(\vect{p}_{t+1}|\vect{z}_{t+1}(\vect{s}_{t+1})\right) \!
    = \! \mathbb{E}_{\vect{z}_{t+1}(\vect{s}_{t+1})} \! \left[   H(\vect{p}_{t+1}|\vect{z}_{t+1}=\vect{z(\vect{s}_{t+1})})\right] \\
    & = \int_{\vect{z}(\vect{s}_{t+1}) \in \mathcal{Z}_{t+1}} p(\vect{z}(\vect{s}_{t+1}))H(\vect{p}_{t+1}|\vect{z}_{t+1}=\vect{z}(\vect{s}_{t+1}))d\vect{z},
\end{split}
\end{equation} 
where, in general, $\mathbb{E}_{\vect{x}} \left[ \cdot \right]$ denotes the expectation w.r.t. the distribution of the $\vect{x}$; $\mathcal{Z}_{t+1}$ is the set of all possible measurements that can be collected at time $t+1$, typically parametrized by the platform state and the control input, from \eqref{eq:sensor_dynamics_general} and \eqref{eq:general_observation_model}. 
In fact, a new measurement is influenced by the next platform state (i.e., $\vect{s}_{t+1}$) and, in turn, by the current platform state (i.e., $\vect{s}_t$) and the actual control input (i.e., $\vect{u}_t$), through \eqref{eq:sensor_dynamics_general}.
In this way, the optimization problem \eqref{eq:input_optimization} is well defined. 
As \eqref{eq:MI_hoffmann2009mobile} suggests, the cost function depends also on the past observations (i.e., $\vect{z}_{1:t}$). From~\eqref{eq:general_observation_model}, each measurement $\vect{z}_k$ is function of the target position; therefore, it can be used to recursively update the target belief map, which is then employed to estimate the target position, or to approximate any function of the target position (e.g., the conditional entropy)~\cite{PF_entropy}.  

If the value of $\vect{z}_{t+1}=\vect{z}$ is known in advance~\cite{ramirez2014moving}, then the conditional entropy would be $H(\vect{p}_{t+1}|\vect{z}_{t+1}=\vect{z})$, which can be computed without taking the  expectation. 
Interestingly, mutual information \eqref{eq:MI_hoffmann2009mobile} is a submodular function~\cite{robotEcology}; thus, solving the optimization problem \eqref{eq:input_optimization} via a greedy algorithm provides an optimality (lower) bound of $63\%$~\cite{krause2011submodularity}. Formally,
\begin{equation}\label{eq:greedy_bound}
    J( \tilde{\vect{u}}_t \;|\; \vect{z}_{1:t},\vect{s}_t) \geq \left( 1-\frac{1}{e} \right) J(\vect{u}_t^* \;|\; \vect{z}_{1:t},\vect{s}_t),
\end{equation}
where $\tilde{\vect{u}}_t$ is the solution of \eqref{eq:input_optimization} obtained with a greedy algorithm.

Information-seeking capabilities can be achieved also by maximizing the expected \textit{Kullback-Liebler} (KL) divergence before and after a new measurement $\vect{z}_{t+1}$ is collected, namely~\cite{hutchinson2019unmanned}
\begin{equation}\label{eq:KL}
\begin{split}
J( \vect{u}_t | \vect{z}_{1:t}, \vect{s}_t) 
\! = \! \mathbb{E}_{\vect{z}_{t+1}} \! \left[ D_{KL} \! \left( \vect{p}_{t+1} | \vect{z}_{1:t},\vect{z}_{t+1}(\vect{s}_{t+1}) ||   \vect{p}_{t+1} | \vect{z}_{1:t} \right) \right] 
\end{split}
\end{equation}
where $D_{KL}(\vect{x}\; || \; \vect{y})$ is the KL divergence between the random vectors $\vect{x}$ and $\vect{y}$~\cite{cover1999elements}. 
The generalization of the KL divergence is the Rényi divergence~\cite{theoretical_sensor_management}; this is parametrized by a parameter $\alpha$, used to give emphasis on specific parts of the distribution. Empirical results~\cite{kreucher2005sensor,aughenbaugh2011sensor} show that the Rényi divergence, for some values of $\alpha \neq 1$, provide superior performance than the KL divergence. It is also possible to prove that maximizing \eqref{eq:KL} and \eqref{eq:MI_hoffmann2009mobile} leads to the same control actions~\cite{theoretical_sensor_management}; in turn, this is equivalent to minimizing the conditional entropy $H(\vect{p}_{t+1}|\vect{z}_{t+1})$, since $H(\vect{p}_{t+1})$ does not depend on $\vect{u}_t$~\cite{haubner2019active,ryan2010particle,robotEcology}. 

The information-theoretic criteria defined so far induce pure explorative behaviors. \textit{Infotaxis}~\cite{infotaxis_nature} is instead used to attain a certain level of
exploration and exploitation trade-off and it is defined according to the following cost function
\begin{equation}\label{eq:infotaxis}
\begin{split}
        J( \vect{u}_t \;|\; \vect{z}_{1:t},\vect{s}_t) 
        & =    p(\vect{p}_{t+1}) H(\vect{p}_{t+1}) 
    \\
    & + \left[ 1-p(\vect{p}_{t+1}) \right] I(\vect{z}_{t+1}(\vect{s}_{t+1});\vect{p}_{t+1}).
\end{split}
\end{equation}
Infotaxis is a gradient-less measure of information gain and it is particularly useful in STE tasks~\cite{park2020cooperative}, due to its capability to work under sparse datapoints.

\subsection{Task-driven control}\label{subsubsec:task_driven}

In literature several definitions of task-driven control have been suggested. A possible task-driven sensor management criterion is based on the minimization of the {\em expected estimation error}; in this case, the cost function employed in the optimization problem~\eqref{eq:input_optimization} is~\cite{theoretical_sensor_management}
\begin{equation}\label{eq:estimation_error}
    J( \vect{u}_t \;|\; \vect{z}_{1:t},\vect{s}_t) =-\mathbb{E}_{\vect{z}_{t+1}(\vect{s}_{t+1})} \left[ e(\vect{p}_{t+1},\hat{\vect{p}}_{t+1}) \right],
\end{equation} 
The estimated target position is denoted as $\hat{\vect{p}}$, while $e(\cdot,\cdot)$ is an estimation performance metric (e.g., Euclidean distance). The definition \eqref{eq:estimation_error} is often impractical, since it may be difficult to find a meaningful metric $e(\cdot,\cdot)$ when there are multiple goals with different importance. The same reason holds when the goals are defined in different dimensions or state spaces~\cite{aoki2011near}. In addition, computing \eqref{eq:estimation_error} incurs a significant computation cost, since it requires to integrate over all possible target positions and all possible sensor measurements~\cite{aoki2011near}. For these reasons, definition \eqref{eq:estimation_error} can be reformulated as~\cite{task_drivenVSinformation_seeking} 
%
\begin{equation}\label{eq:expected_MAP}
\begin{split}
        & J( \vect{u}_t \;|\; \vect{z}_{1:t},\vect{s}_t) \! = \! \mathbb{E}_{\vect{z}_{t+1}(\vect{s}_{t+1})} \! \left[ J^\prime( \vect{u}_t \;|\; \vect{z}_{1:t},\vect{s}_t)  \right],\\
        &  J^\prime( \vect{u}_t \;|\; \vect{z}_{1:t},\vect{s}_t) = \max_{\vect{p}_{t+1} \in \Pi} p(\vect{p}_{t+1} |\vect{z}_{1:t}, \vect{z}_{t+1}(\vect{s}_{t+1}))
\end{split}
\end{equation}
%
which leads the platform to take the action that leads to the largest Maximum A-Posteriori ({\em MAP}) estimate of target location, that is the action that results in maximal certainty about target position. 

In target search applications~\cite{liu2015model}, a commonly used task-driven criterion is the maximization of the {\em expected POD} over a given time horizon $\nu$, that is~\cite{liu2015model}
\begin{equation}\label{eq:POD_criterion}
\begin{split}
    J( \vect{u}_t | \vect{z}_{1:t},\vect{s}_t) 
    & = \mathbb{E}_{\vect{p}_{t+1}} \! \left[ p \left( \bigcup_{k=t+1}^{t+\nu} D_k(\vect{s}_{t+1}) \left \vert{ D_{1:t} = \emptyset, \vect{z}_{1:t} }\right. \right) \! \right] \\
    & \!=\! - \mathbb{E}_{\vect{p}_{t+1}} \! \left[ \prod_{k=t+1}^{t+\nu} p \left( \bar{D}_k(\vect{s}_{t+1})  \! \left \vert{ D_{1:t} \!=\! \emptyset, \vect{z}_{1:t} }\right. \right) \! \right].
\end{split}
\end{equation}
In \eqref{eq:POD_criterion},  $\bar{D}_{t}$ denotes a non-detection event at time $t$ and $p\left( D_t | \bar{D}_{1:t-1} \right)$ is the probability of detecting the target at instant $t$, given the non-detection in previous observations. As can be seen from the second equation, each detection event is considered independent to the others.
Traditionally, the calculation of \eqref{eq:POD_criterion} incurs high computation burden because no analytical form exists and numerical integration over large state space is required. Thus, ad-hoc approximation techniques have been proposed to efficiently solve POD-based search tasks also in real-time applications. For instance, by approximating the target belief map with a Gaussian Mixture Model (GMM)~\cite{liu2015model} it is possible to get a closed-form reformulation of \eqref{eq:POD_criterion}. Alternatively, it is possible to assume the detection probability as constant during the time horizon $[t+1,t+\nu]$; in this way, the target search problem can be mathematically modeled as a Binomial experiment.

{\em Minimum Time Search} (MTS)~\cite{MTS,MTS_Rinner,lanillos2012minimum} is a task-driven criterion closely related to the POD function. In fact, it aims at minimizing the expected detection time, where the time of first detection ($T_D$) is modeled as a random variable; thus, the cost function in this case is~\cite{lanillos2012minimum,MTS,perez2018ant}
\begin{equation}\label{eq:ET}
\begin{split}
    J( \vect{u}_t \;|\; \vect{z}_{1:t},\vect{s}_t) 
    & = - \mathbb{E} \left[ T_D \right] \\
    & = - \sum_{k=t+1}^{\infty} \left[ 1 - p( T_D \leq k )\right]\\
    & = - \sum_{k=t+1}^{\infty} k \; p\left( D_k(\vect{s}_{t+1}) | \bar{D}_{1:k-1}, \vect{z}_{1:t}\right) 
\end{split}
\end{equation}
The definition in \eqref{eq:ET} is intractable; for this reason, a truncated time horizon is often considered in real-life applications~\cite{lanillos2012minimum,lanillos2013bayesian,perez2018ant}.
The main difference between the POD criterion \eqref{eq:POD_criterion} and the MTS one \eqref{eq:ET} stems from the fact that the former maximizes the chances of finding the target, but it does not guarantee to minimize the time to first detection; therefore, the POD criterion induces a slightly more explorative behavior during the search mission~\cite{lanillos2012minimum}.

{\em Source seeking} is another example of task-driven control, often accomplished via probabilistic approaches. In particular, a target-representative setpoint (also known as reference point) is extracted from the belief map; then, the platform is driven towards the setpoint. The MAP~\cite{radak2017moving} or the MMSE\footnote{MMSE: Minimum Mean Square Error.}~\cite{hasanzade2018rf} target position estimates are typical setpoint choices. In source seeking, the cost function is the distance between the current platform position\footnote{When the platform position is fixed and only rotation movements are allowed (e.g., pan cameras on a static robot), the distance is computed between the setpoint and the centre of the FoV projection onto the groundplane $\Pi$~\cite{chen2014novel}} and the setpoint itself, namely
\begin{equation}\label{eq:source_seeking}
\begin{split}
    J( \vect{u}_t \;|\; \vect{z}_{1:t},\vect{s}_t) =
    - \lVert \vect{s}_t - \hat{\vect{p}}_t \rVert_2.
\end{split}
\end{equation}
This strategy is motivated by the assumption that moving closer to the target reduces the uncertainty of measurements~\cite{van2020lavapilot}; however, it becomes inefficient if the belief map is characterized by multiple peaks~\cite{radmard2017active}. 
Finally, when time of search matters, source seeking can be combined with MTS~\cite{liu2015model}. 

\vspace{0.2cm}

To sum up, there is a certain lack of consistency, among the reviewed works, regarding the definition of task-driven control. 
A task-driven APE criterion can be simply defined as the opposite of an information-seeking criterion, that is a control law is defined as task-driven when the cost function is not information-theoretic and drives the platform closer to its task completion in an exploitative manner. 
For example, the  goal position can be reached within certain time/energy constraints, minimizing the estimation error and maximizing the probability of target detection. 

\subsection{Hybrid control}\label{subsubsec:hybrid}

Many practical situations require a trade-off between exploration and exploitation~\cite{gupta2006interplay,xu2014exploration,meera2019obstacle}, as most APE tasks are designed to accurately reconstruct information about the target location, but not to produce overly time-consuming trajectories during the mission~\cite{vander2015algorithms}.

For this reason, some works propose hybrid control techniques that balance information-seeking and task completion. 
To do this, the most intuitive way is to cast the problem \eqref{eq:input_optimization} into a {\em multi-objective optimization} framework~\cite{marler2004survey}, where the cost function is a linear combination between an information gathering measure and a task execution metric~\cite{liu2017model,ramirez2014moving}. 

Alternative approaches cast the APE problem into a BO framework~\cite{frazier2018tutorial,snoek2012practical,meera2019obstacle}, which is a powerful strategy for finding the extrema of an objective function in a gradient-free manner and when the cost function is expensive to evaluate, noisy, or non-convex~\cite{brochu2010tutorial}. 
In BO, the objective function is approximated by a surrogate function (e.g., GP~\cite{rasmussen2003gaussian}); then, an acquisition function (e.g., Gaussian Confidence Upper Bound~\cite{BayOpt_exploration}, Expected Improvement~\cite{carpin2015uavs}) chooses the next sensing action, usually managing the trade-off between exploration and exploitation~\cite{meera2019obstacle}. The acquisition function can also be a linear combination  between  a  source  seeking  term  and  an  information gathering one~\cite{ghassemi2020extended}; the former might account for the distance to  the  maximum  expected  target  position,  while  the  latter should depend on the target belief uncertainty.

Instead of combining exploration and exploitation in a single cost function, it is also possible to use \textit{switching controllers} to alternate between an information-seeking contribute (e.g., predictive variance) and a source-seeking one (e.g., distance to the Maximum Likelihood target estimate)~\cite{fink2010online}. In any case, switching controllers are liable to unstable behaviors; hence, when they are used, strict convergence guarantees should be provided. 

\subsection{Discussion}


According to some previous works~\cite{task_drivenVSinformation_seeking}, task-driven control slightly outperforms information-seeking in terms of tracking error; in fact, information-based approaches sacrifice a small amount of tracking performance (i.e., estimation bias) to yield smaller estimation uncertainty. In general, explorative strategies are used when the priority is on uncertainty reduction, or if the APE task includes multiple objectives. On the contrary, exploitative strategies are specifically designed for specialized tasks (e.g., source seeking, detection time minimization). This usually requires ad-hoc utility functions, which is the main limitation of task-driven controllers, as wrong choices may lead to local minima, as well as to unstable and ineffective platform behaviors~\cite{radmard2017active}. 

\subsubsection{Challenges in task-driven control}
To increase the flexibility of pure-exploitative strategies, hybrid solutions can be adopted, even though 
in hybrid control the performance strongly depends on accurate exploitation-exploration balancing methods~\cite{ghassemi2020extended}: if the controller is biased towards the source-seeking term, the estimation process might get stuck on local optima and the target may never be found; on the other hand, higher importance on the exploration term, leads to superior capabilities to explore the environment and, therefore, to find the target. However this may come at the cost of higher completion times, sometimes not compatible with the mission requirements~\cite{ghassemi2020extended}. In conclusion, hyperparameter tuning is not only cumbersome and time-consuming, but also critical for the performance of hybrid APE solutions. For this reason, automatic hyperparameter tuning algorithms are currently an active research field in machine learning and robotics communities.

\subsubsection{Challenges in information-seeking control}
While pure-exploitative strategies lack in flexibility, one of the main pitfalls of information-seeking approaches is the poor interpretability~\cite{aoki2011near}, together with a limited capability in considering time and energy preservation policies. Furthermore, unlike task-based algorithms, information-seeking path planning is not adequate to real-time APE missions, where platforms have limited energy and computation power~\cite{van2020lavapilot}. This is due to the intractability of the numerical integration methods
involved in the calculation of most information-theoretic cost functions (e.g., KL divergence, entropy, information gain)~\cite{hoffmann2009mobile}. To reduce the computational burden, parametric~\cite{liu2015model} and non-parametric~\cite{ryan2010particle,realTime_active_vision,aughenbaugh2011sensor} approximations are often applied; these, however, have an impact on the quality of the overall solution~\cite{hoffmann2009mobile} and restrict the range of techniques that can be used to solve the optimization problem \eqref{eq:input_optimization}. In particular, the lack of any analytical closed-form solution leads to the use of grid-search~\cite{ryan2010particle}, greedy~\cite{robotEcology} and meta-heuristic~\cite{MTS_Rinner} techniques, which are not always efficient and accurate~\cite{ramirez2014moving,liu2015model}. 
%

\subsubsection{Relations between task-driven and information-seeking strategies}
The {\em near universal proxy} theory directly links any arbitrary task-driven function to the optimization of a criterion based on the Rényi divergence~\cite{hero2008information}. In particular, the  near universal proxy theory  claims that the expected value of any task-driven function is sandwiched between two marginalized Rènyi divergences; consequently, task-driven sensor management criteria could be replaced by a Rènyi divergence-based criterion without altering the performance guarantees.
Even though the near universal proxy argument is so far the strongest theoretical justification for the use of information-driven sensor management, it has been formally rebutted~\cite{aoki2011near}; hence,
there is still lack of a solid proof regarding the relationship between task-driven and information-driven sensor management.

On the opposite side of the near universal proxy theory, some works~\cite{ryan2008information,optimal_information_search} have proven that it is possible to reformulate an information-theoretic optimization problem into an equivalent task-driven one. This, in turn, is less costly and can sometimes be calculated more accurately; hence, reformulating information-seeking tasks into task-driven ones can solve the computational issues discussed above. The main argument against these results is that they usually hold under specific conditions. For instance, the minimization of the expected entropy is equivalent to the maximization of the POD, if the POD is sufficiently small~\cite{ryan2008information}. If instead the POD is constant, the optimization problem is equivalent to the maximization of the expected KL divergence (in the sense that both generate the same control input)~\cite{optimal_information_search}. Note however that the POD is a joint function of the target state and the control input, as outlined in Sec. \ref{sec:APE}; hence, supposing this quantity to be constant over the entire $\Pi \times \mathcal{S}_m$ space is an ideal assumption, difficult to be met in real-life scenarios (see Fig. \ref{fig:camera_fov}). 

\subsubsection{Future research lines}
In spite of the presence of some comparative studies~\cite{optimal_information_search,task_drivenVSinformation_seeking,theoretical_sensor_management,van2020lavapilot}, the relation between task-driven and information-seeking approaches is still an open debate in APE-related research fields. Future research should aim to develop a generalized and unified guideline for the choice of the APE objective function to be used in \eqref{eq:input_optimization}. This should also account for the environment conditions, the number of cooperating platforms and their available resources (e.g., on-board computational and power capabilities), the platforms constraints, and for any specific task requirement (e.g., minimum time search). 
Another possible future trend in APE literature is the design of {\em Minimum Energy Search} (MES) strategies, which is still an unexplored line of research. Albeit MES might share some similarities with MTS, it is important not to confuse the two approaches: as discussed in Sec. \ref{subsec:criteria}, MTS generates control inputs that minimize the expected detection time \eqref{eq:ET}, but no guarantees are provided on the platform energy-consumption, which is an important limitation in real-life applications.


\section{Applications}
\label{sec:applications}

Most real-life scenarios involve the coexistence of different APE requirements; hence, in practice, the transition between each specific APE class is not always clear and well distinguishable. To highlight this coexistence of requirements, in this Section we provide an overview of representative APE applications, including STE, wildlife monitoring, SAR missions, multi-robot cooperative systems and collaborative robotics. 
%


AS is widely used in the field of SAR missions, where aerial~\cite{SAR_evolutionary} or ground~\cite{rodriguez2020wilderness} robots, embedded with visual~\cite{SAR_UAV}, acoustic~\cite{sibanyoni20182} or radio~\cite{SAR_radio} devices, are employed to detect~\cite{radmard2017active} and localize~\cite{stone2011search} people in imminent danger. 
Accidental releases of liquid or gaseous pollutants (i.e., HAZMAT) into the ambient environment can occur at industrial facilities such as petroleum refineries, chemical plants, oil and gas transportation pipelines, and many other industrial activities. 
HAZMAT pose both an immediate and chronic risk to environment and human health; hence, a prompt and accurate STE is important to enable appropriate mitigation steps. In particular, autonomous STE allows to perform the task without endangering the personnel. In this case, the target is the HAZMAT source (usually static), while the platforms are typically unmanned ground or aerial vehicles~\cite{STE_review}, equipped with specific HAZMAT detectors~\cite{STE_review} (e.g, atmospheric sensors~\cite{hutchinson2019unmanned}). According to the concentration of the HAZMAT, it is possible to retrieve the source position~\cite{park2020cooperative} by leveraging on specific dispersion models~\cite{STE_review}.

\textit{Wildlife monitoring} is essential for understanding animal movement patterns, habitat utilisation, population demographics and poaching incidents. Radio transmitters are usually attached to animals, so that autonomous robotic platforms (e.g., UAVs) embedded with radio receivers can track their movements~\cite{robotEcology}. Automated wildlife tracking is necessary because robot systems can access rugged areas that are difficult for humans to traverse; moreover, manually locating radio transmitters with handheld
equipment is labor-intensive. 

The recent development in aerial mobile robotics~\cite{DJI_autoReturn} has revolutionized several applications that require dynamic camera viewpoints, such as \textit{autonomous cinematography}~\cite{jeon2020detection,bonatti2019towards} and \textit{entertainment}~\cite{kim2018survey}. In films production, flying agents are employed to follow actors motion, moving through an unknown environment at
high speeds. At the same time, they are required to map the environment to provide artistic and technical feedback to the producers. In this contexts, to plan smooth, collision-free trajectories while avoiding occlusions, is extremely challenging even for experienced pilots. For this reason, the current trend is to exploit autonomous active tracking modules~\cite{bonatti2019towards}. 

Localization and tracking is fundamental in multi-robot cooperative systems~\cite{nagaty2015probabilistic,haugen2015monitoring,carron2015multi,rizk2019cooperative}, where teams of robots work together to complete a mission~\cite{aranda2015formation,chen2015coordination,lissandrini2019cooperative}. 
In \textit{autonomous warehouse} inventory schemes~\cite{guerin2016towards}, UGVs carries bring aerial platforms among rows of racks; then, the UAVs are used to scan or pick goods that are not accessible to the ground robot. From the UGV perspective, the objective is to locate and reach a desired warehouse planar position. On the other hand, the target of the UAV is a specific good, after the take-off, and the UGV carrier during landing. Both tasks must be accomplished in an autonomous manner, relying on APE strategies based visual, acoustic or radio signals.  

Another application of multi-robot cooperation is rendezvous planning for \textit{mobile recharging} stations~\cite{recharge,yu2018algorithms}; it consists of a cooperative replenishment strategy for a team of working robots (UAVs) performing a surveillance task, using one or more mobile charging robots. Each charging robot is equipped with a payload of batteries and automated battery swap systems, and the goal is to schedule paths for charging robots such that they rendezvous with and replenish the UAVs, as needed, during the mission.

\textit{Collaborative robotics}~\cite{vicentini2021collaborative} aims at designing robotic platforms capable of cooperating with human beings. Most applications in this field require to actively search, localize and track people by using onboard sensors. To this aim, novel technologies have been proposed to extract contextual information regarding user behaviors (e.g., health monitoring of elderly people, gesture recognition for smart home applications)~\cite{jiang2018smart}; this data can be used to activate positioning tasks in the context of \textit{assisted living} and \textit{ambient intelligence}~\cite{sokullu2019role}. For instance, if a fall detection system~\cite{jiang2018smart} raises a health monitoring alarm, then a mobile platform might be used to search for the user and visually verify their conditions~\cite{lewis2016framework}.

\section{Conclusion}
\label{sec:conclusion}

In this paper, we organized, critically discussed and compared APE methods, with a focus on perception, control and multi-agent cooperation. We defined a taxonomy and a formal problem statement, and focused on active and probabilistic approaches. 
We also provided a formal characterization of search and tracking tasks, and covered multi-modal (both single-platform and multi-platform) APE strategies, with an exhaustive analysis of multi-agent cooperative algorithms. 
Furthermore, we compared information-seeking and task-driven APE criteria, and 
comprehensively discussed key decision-making and path-planning criteria. 

Future research
directions include working on \emph{multi-modal} approaches and the determination of how different sensing modalities impact SAT tasks as well as the comparison among different APE criteria, as the near-universal proxy argument is still an open problem. Moreover, it is still unclear how to balance exploration with exploitation, especially under critical mission constraints. To this aim, we believe that  the concept of \emph{hybrid control} for APE missions that we introduced in this survey is the most promising research direction for robotics applications. 
%






\bibliographystyle{IEEEtran}
\bibliography{IEEEabrv,References}

\begin{IEEEbiography}
[{\includegraphics[width=1in,height=1.25in,clip,keepaspectratio]{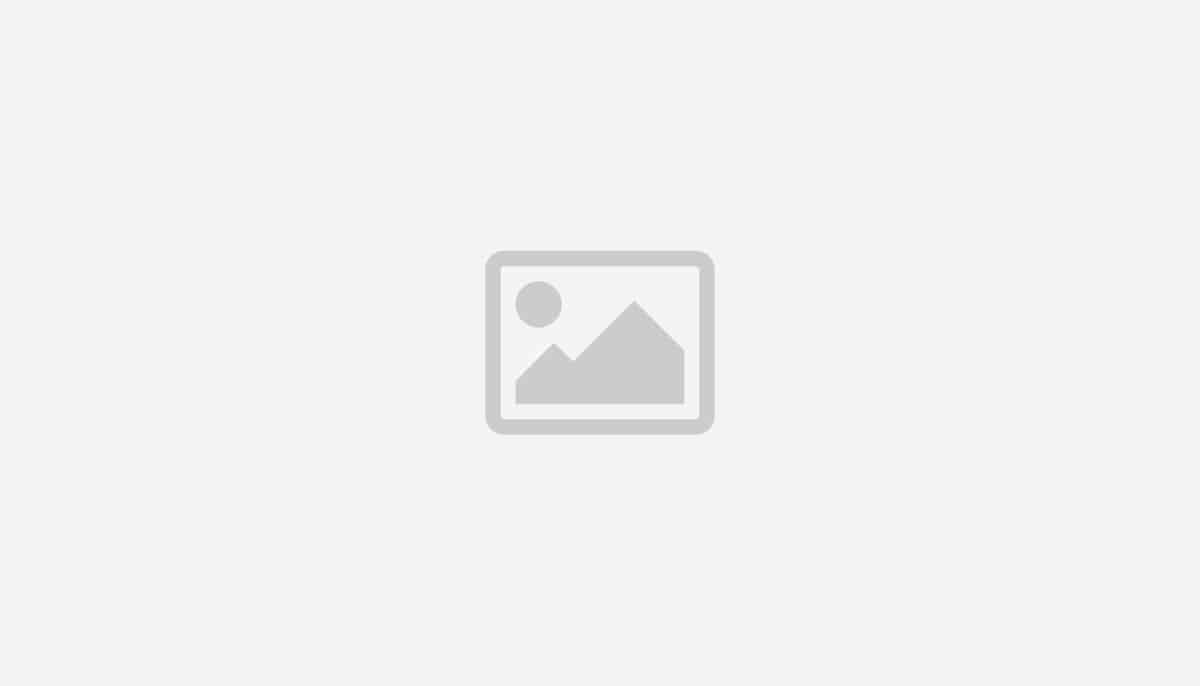}}]{Name Surname} ... .
\end{IEEEbiography}

\end{document}